\documentclass[runningheads]{llncs}

\usepackage{eccv}

\usepackage{eccvabbrv}
\usepackage[table]{xcolor}
\usepackage{graphicx}
\usepackage{booktabs}
\usepackage{multirow}
\usepackage{wrapfig}

\usepackage{hyperref}

\usepackage{tabularx}
\usepackage{array}
\usepackage{makecell}
\usepackage{amssymb} 
\newcolumntype{C}{>{\centering\arraybackslash}X}

\newcommand{\Yes}{\ensuremath{\checkmark}}
\newcommand{\No}{\ensuremath{\times}}
\usepackage[most]{tcolorbox}
\tcbset{colback=gray!5!white,colframe=gray!40!white,boxrule=0.4pt,arc=1mm,left=2mm,right=2mm,top=1mm,bottom=1mm}

\newtcblisting{promptbox}[1]{
  listing only,
  listing options={
    basicstyle=\ttfamily\footnotesize,
    breaklines=true,
    breakindent=0pt,
    columns=fullflexible
  },
  breakable,
  title={#1},
  fonttitle=\bfseries\small,
  coltitle=black
}
\begin{document}

\title{EpiBench: Benchmarking Multi-turn Research Workflows for Multimodal Agents} 


\author{Xuan Dong$^{*}$\inst{1} \and
Huanyang Zheng$^{*}$\inst{1} \and
Tianhao Niu$^{*}$\inst{1} \and
Zhe Han\inst{1} \and
Pengzhan Li\inst{1} \and
Bofei Liu\inst{1} \and
Zhengyang Liu\inst{1} \and
Guancheng Li\inst{1} \and
Qingfu Zhu$^{\dagger}$\inst{1} \and
Wanxiang Che$^{\dagger}$\inst{1}}

\authorrunning{X. Dong et al.}

\institute{Harbin Institute of Technology\\
\email{\{qfzhu,car\}@ir.hit.edu.cn}\\
$^{*}$Equal contribution\\
$^{\dagger}$Corresponding authors}

\maketitle

\begin{abstract}
  Scientific research follows multi-turn, multi-step workflows that require proactively searching the literature, consulting figures and tables, and integrating evidence across papers to align experimental settings and support reproducible conclusions. This joint capability is not systematically assessed in existing benchmarks, which largely under-evaluate proactive search, multi-evidence integration and sustained evidence use over time. In this work, we introduce EpiBench, an episodic multi-turn multimodal benchmark that instantiates short research workflows. Given a research task, agents must navigate across papers over multiple turns, align evidence from figures and tables, and use the accumulated evidence in the memory to answer objective questions that require cross paper comparisons and multi-figure integration. \textsc{EpiBench} introduces a process-level evaluation framework for fine-grained testing and diagnosis of research agents. Our experiments show that even the leading model achieves an accuracy of only 29.23\% on the hard split, indicating substantial room for improvement in multi-turn, multi-evidence research workflows, providing an evaluation platform for verifiable and reproducible research agents.

\end{abstract}

\section{Introduction}
\label{sec:intro}
The accelerating expansion of the scientific literature has made literature review, verification, and experimental reproduction increasingly costly and time consuming, creating a pressing demand for automated research assistance \cite{vishesh2025aegisagentextractiongeographic,10.1145/3747912.3747962}.
This demand has fueled growing interest in research agents for literature review, verification, and experimental reproduction. Recent advances in large multimodal models and tool-using agents make this direction increasingly practical, enabling systems to browse the web, read PDFs, invoke external tools, and produce citation-supported answers with intermediate traces ~\cite{yao2023react,NEURIPS2023_d842425e,qin2023toolllmfacilitatinglargelanguage}. As a result, agents are now expected to locate relevant articles, interpret figures and tables, align experimental settings across sources, and deliver conclusions that can be checked and reproduced.

\begin{table*}[t]
\caption{Comparison of agentic reasoning benchmarks by key capabilities. \Yes\ indicates supported, and \No\ indicates not supported.}
\label{tab:benchmark_compare}
\centering
\scriptsize
\setlength{\tabcolsep}{0pt}
\renewcommand{\arraystretch}{1.08}
\begin{tabularx}{\textwidth}{lCCCCC}
\toprule
\textbf{Benchmark} &
Cross-paper Integration &
Multimodal Integration &
Proactive Search &
Evidence Memory &
Multi-turn Dialogue \\
\midrule

\multicolumn{6}{c}{\textit{Non-paper benchmarks}} \\
\midrule
MMCR\cite{yan2025mmcradvancingvisuallanguage} & \No & \No & \No & \No & \Yes \\
Vision-DeepResearch\cite{zeng2026visiondeepresearchbenchmarkrethinkingvisual} & \No & \Yes & \Yes & \No & \No \\
MMDU\cite{liu2024mmdumultiturnmultiimagedialog} & \No & \Yes & \No & \No & \Yes \\
VisChainBench\cite{lyu2025vischainbenchbenchmarkmultiturnmultiimage} & \No & \Yes & \No & \No & \No \\
MMDeepResearch\cite{huang2026mmdeepresearchbenchbenchmarkmultimodaldeep} & \Yes & \Yes & \Yes & \No & \No \\
MMSearch-Plus\cite{tao2025mmsearchplusbenchmarkingprovenanceawaresearch}  & \No & \Yes & \No & \No & \No \\
\midrule

\multicolumn{6}{c}{\textit{Paper-centric benchmarks}} \\
\midrule
LiveXiv\cite{shabtay2025livexivmultimodallive}  & \No & \No & \No & \No & \No \\
CharXiV\cite{wang2024charxivchartinggapsrealistic} & \No & \No & \No & \No & \No \\
SPIQA\cite{pramanick2025spiqadatasetmultimodalquestion} & \No & \No & \No & \No & \No \\
SciVQA\cite{movva2025enhancingscientificvisualquestion} & \No & \No & \No & \No & \No \\
SCIGraphQA\cite{li2023scigraphqa} & \No & \No & \No & \No & \No \\
SIN-Bench\cite{ren2026sinbenchtracingnativeevidence} & \No & \Yes & \No & \No & \No \\
PaperArena\cite{wang2026paperarenaevaluationbenchmarktoolaugmented} & \Yes & \Yes & \No & \No & \No \\
\textbf{EpiBench (Ours)} & \Yes & \Yes & \Yes & \Yes & \Yes \\
\bottomrule
\end{tabularx}
\end{table*}

Despite these advances, current evaluations still fall short of real research workflows, especially in paper-related settings. Table~\ref{tab:benchmark_compare} suggests two fundamental mismatches. First, \textbf{benchmark design} is often not workflow-faithful. 
Many tasks begin from an explicitly specified target paper or direct identifiers, so agents are not required to perform \textit{proactive search} from citation cues or indirect hints.
When the target is already known, questions are frequently answerable from a single figure or table, and \textit{cross-paper multimodal integration} over \textit{multiple figures and tables} is seldom necessary \cite{ren2026sinbenchtracingnativeevidence,wang2024charxivchartinggapsrealistic}.
In addition, many benchmarks provide limited coverage of \textit{multi-turn} research episodes where evidence must be accumulated, refined, and reused over time. Second, the \textbf{evaluation protocol} is incomplete.
Because repeated re-browsing is typically allowed, benchmarks rarely measure whether an agent \textit{reuses previously acquired evidence}, and they seldom check whether intermediate evidence is correctly grounded, which makes evidence attribution errors and integration failures hard to diagnose \cite{shabtay2025livexivmultimodallive,pramanick2025spiqadatasetmultimodalquestion}. 
These two mismatches naturally lead to two core gaps. \textit{(Problem~1)} existing benchmarks do not comprehensively assess human-like workflows that require \textit{proactive search}, \textit{multi-turn interaction}, and \textit{cross-paper multimodal evidence fusion} over time. \textit{(Problem~2)} existing protocols lack process-level metrics for \textit{evidence reuse} and evidence correctness needed for fine-grained diagnosis.


\begin{figure*}[t]
  \centering
  \includegraphics[width=\textwidth]{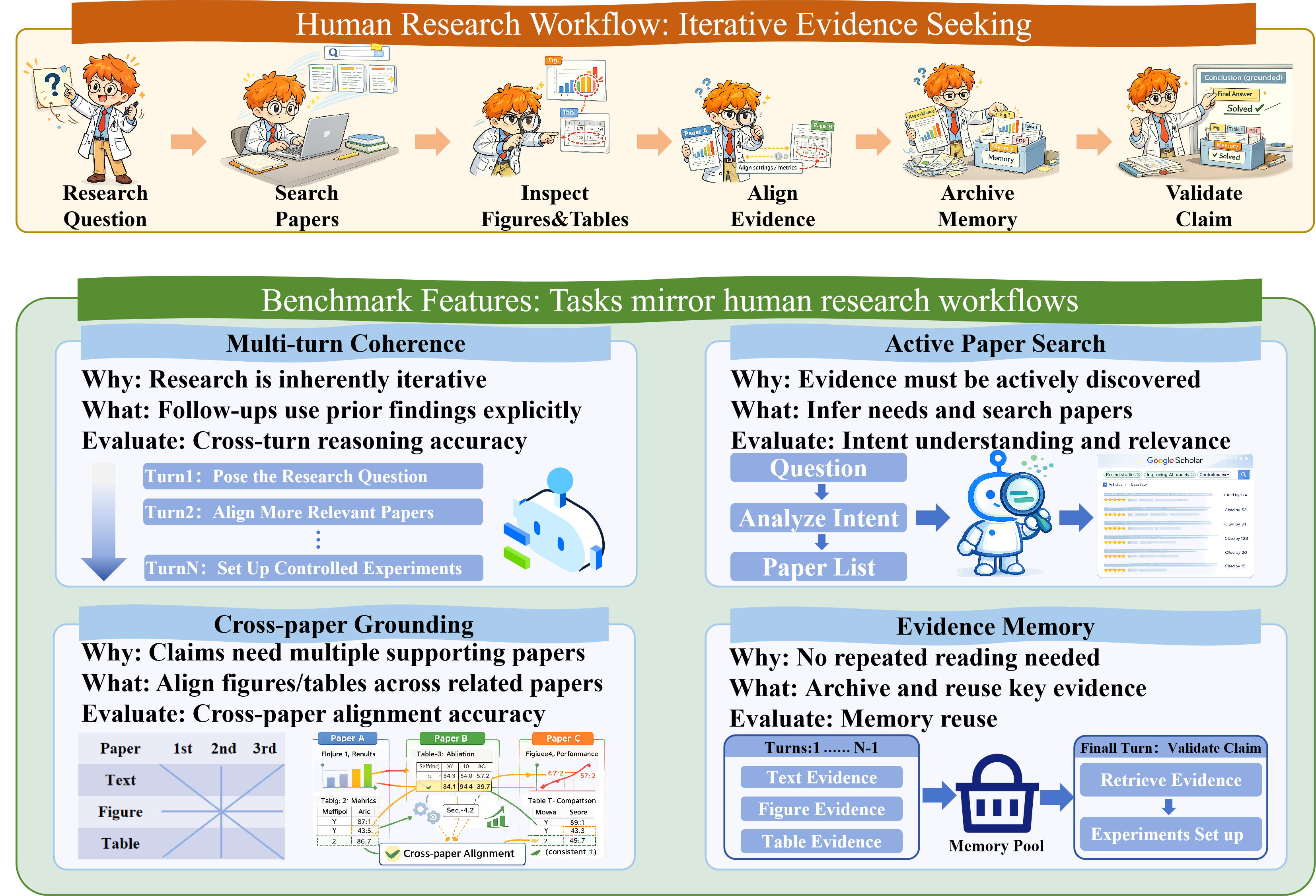}
  \caption{\textbf{EpiBench motivation and key features.} \textbf{Top:} A human research workflow proceeds iteratively from posing a question to searching papers, inspecting figures and tables, aligning evidence across papers, archiving findings, and validating the final claim. \textbf{Bottom:} \textsc{EpiBench} mirrors this workflow with four benchmark features: multi-turn coherence, active paper search, cross-paper grounding over figures and tables, and evidence memory that supports reuse in later turns.}
  \label{fig:workflow_features}
\end{figure*}

To address \textit{Problem~1}, we introduce \textsc{EpiBench}, a benchmark that evaluates research workflows at the process level. Each instance is an episodic multi-turn task that simulates a short research process. In our formulation, the task description does not provide a target paper or direct identifiers, mirroring realistic user queries that rarely include precise citations. This setting requires agents to perform \textbf{proactive search}, using \textbf{citation cues} and \textbf{indirect hints} to discover \textbf{multiple papers}, extract structured evidence from specific \textbf{figure and table regions}, and repeatedly integrate evidence across turns to produce the final answer. The questions are objective and enforce cross-paper multi-figure and multi-table alignment through comparisons, selection rules, and set-style constraints grounded in visual legends and plotted values. This design reduces the possibility of answering from prior world knowledge or incomplete visual context, and it directly tests an agent's ability to accumulate and integrate multimodal evidence over multi-turn interaction.

To address \textit{Problem~2}, \textsc{EpiBench} goes beyond task design and introduces an access-budget evaluation setting that quantifies evidence reuse and efficiency from agent interaction logs.
As illustrated in Fig.~\ref{fig:workflow_features}, we focus on human-like research workflows where evidence is actively collected, aligned, and reused across turns. 
To enforce memory-based reuse, we restrict the agent to answering questions exclusively from memory without invoking external tools when reuse is required.
To measure reuse correctness and evidence grounding, we annotate required evidence units at the granularity of paper identifiers and figure or table identifiers, enabling automatic provenance-based scoring. 
Using these traces, we evaluate not only final correctness but also evidence reuse and tool efficiency, and we diagnose failures such as retrieval errors, perception errors, stale or inconsistent reuse, and reasoning errors.

We conduct extensive experiments with representative open source and closed source Large Multimodal Model (LMM) based agents and observe that research workflow performance remains limited. Beyond the final accuracy, intermediate traces reveal brittle behavior. 
Agents often rely on incomplete visual evidence, link answers to incorrect figures or tables, and resort to erroneous memory or prior knowledge instead of seeking information that truly supports the evidence.
These failure modes are difficult to diagnose with accuracy alone, but are exposed by \textsc{EpiBench} through process level evaluation of evidence grounding, memory reuse, and tool efficiency. 
We expect \textsc{EpiBench} to support the development of more reliable research agents and to accelerate progress in multimodal literature understanding and evidence based scientific reasoning.

Our main contributions are summarized as:
\begin{itemize}
\item We introduce \textsc{EpiBench}, an episodic workflow-level benchmark for research agents, filling a key gap in prior evaluations that largely lack process-level research-workflow tasks. It requires proactive search across papers and joint use of figures, tables, and text, with each retrieved paper contributing essential evidence to the final answer.
\item We propose a budgeted evaluation protocol with annotations and diagnostics of the evidence unit to measure episode success, evidence grounding, evidence reuse, and tool efficiency under explicit access constraints of paper and figures.
\item We identify robustness weaknesses of current LMM agents on research-workflow tasks: they often hallucinate answers based on prior knowledge or incomplete visual evidence, and fail to reliably retrieve, ground, and reuse evidence correctly across papers and figures over multiple turns.
\end{itemize}

\section{Related Work}

\subsection{Benchmarks on Scientific Literature}

Recent benchmarks have explored scientific document understanding, particularly focusing on figures and charts in scholarly papers. 
Datasets such as SciVQA\cite{movva2025enhancingscientificvisualquestion}, CharXiv\cite{wang2024charxivchartinggapsrealistic}, and SPIQA\cite{pramanick2025spiqadatasetmultimodalquestion} study question answering over scientific figures, while LiveXiv\cite{shabtay2025livexivmultimodallive} extends this paradigm by generating large-scale and continuously updated VQA tasks from academic papers. 
More recent work has focused on multimodal reasoning across figures and textual context. 
For example, SIN-Bench\cite{ren2026sinbenchtracingnativeevidence} simulates multi-step workflows such as evidence discovery and grounded question answering in long multimodal documents.
PaperArena~\cite{wang2026paperarenaevaluationbenchmarktoolaugmented} is a closely related benchmark that evaluates tool-augmented multi-hop navigation across academic papers. However, its tasks rarely require jointly integrating information from multiple figures and tables to derive an answer.

However, these benchmarks remain largely limited to single-document settings. Most tasks involve answering isolated questions about a single figure, table, or text span within one paper, so they rarely require integrating multimodal evidence across multiple figures and tables or aligning evidence across multiple papers, and they seldom require reusing information obtained from earlier questions. In contrast, \textsc{EpiBench} evaluates agents in a setting that requires cross-paper reasoning, multi-figure and multi-table evidence integration, and iterative evidence accumulation and reuse across turns.

\subsection{Multi-Turn Benchmarks and Long-Horizon Memory}

Several recent benchmarks investigate multi-turn reasoning in multimodal settings. MMDU\cite{liu2024mmdumultiturnmultiimagedialog} introduces multi-image, multi-turn dialogues derived from Wikipedia content. 
MULTIVERSE\cite{lee2025multiversemultiturnconversationbenchmark} further expands this direction by constructing a dataset with greater scale and depth, while MMCR emphasizes contextual reasoning across dialogue turns to improve model performance.

The development of multi-turn benchmarks has extended the reasoning horizon required to solve a question, necessitating that agents reuse information accumulated during earlier reasoning steps. However, existing studies (e.g., Memory-QA~\cite{jiang2025memoryqaansweringrecallquestions} and EMP~\cite{chen2026efficientmultimodalplanningagent}) primarily focus on retrieving answers to individual queries from previously stored memories. Moreover, they are generally not paper-centric and rarely require cross-paper multimodal evidence integration over figures and tables. In contrast, \textsc{EpiBench} targets research workflows by requiring agents to iteratively explore multiple papers, integrate multimodal evidence, and build upon prior findings across turns to answer objective, evidence-grounded scientific questions.

\section{EpiBench: the benchmark}
In this section, we introduce \textsc{EpiBench} and describe its benchmark design. We first formalize the episodic task setting and the corresponding evidence and process annotations, followed by our benchmark construction pipeline. Furthermore, we summarize the key features that distinguish \textsc{EpiBench} from existing benchmarks. Finally, we define the evaluation metrics used throughout the paper.
\subsection{Task Formulation}
\label{sec:task}

\begin{figure*}[t]
  \centering
  \includegraphics[width=\textwidth]{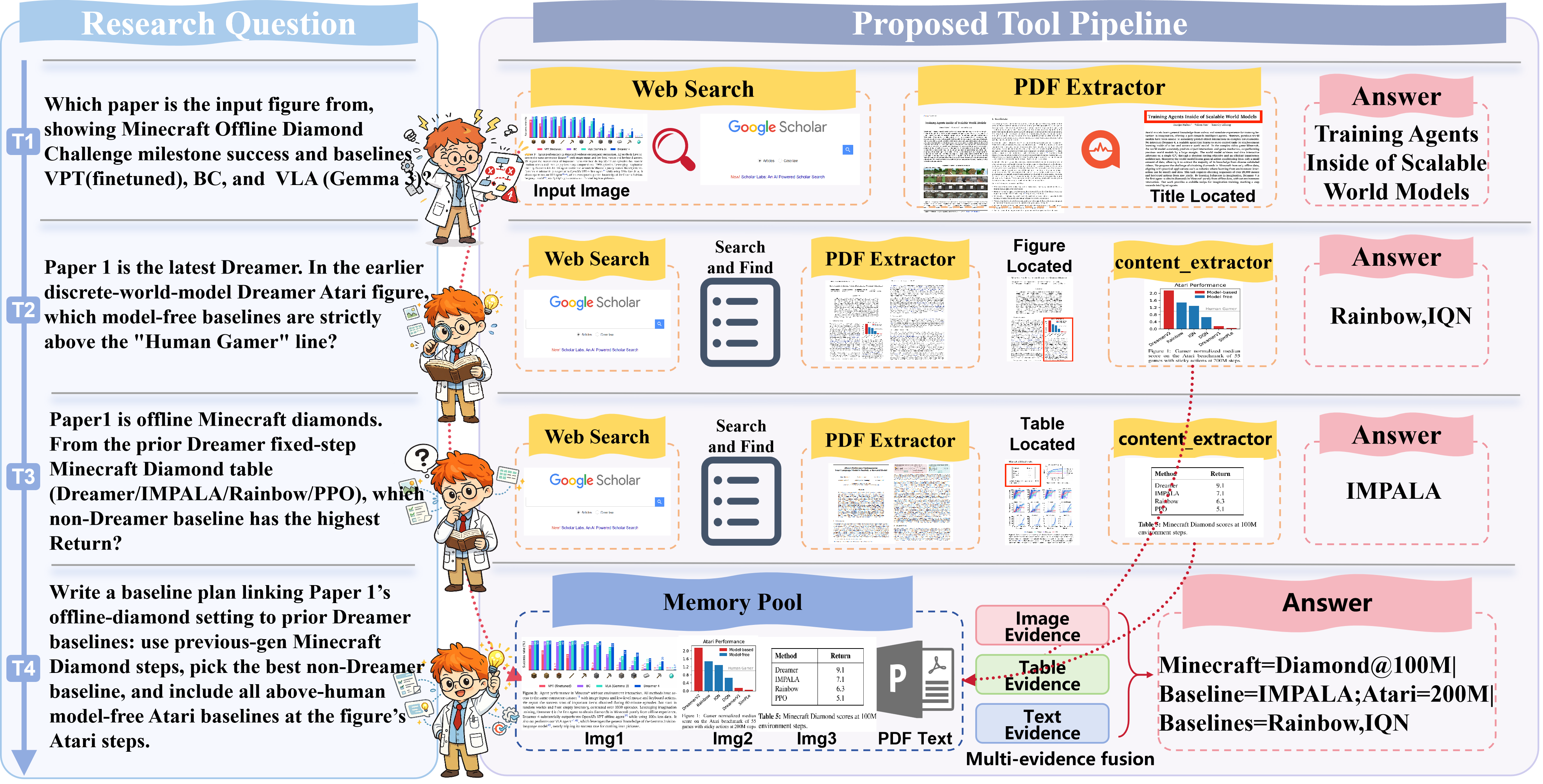}
  \caption{\textbf{Example episode and tool-use workflow in \textsc{EpiBench}.} A research task is instantiated as a multi-turn question chain. At each turn, the agent uses tools to search for relevant papers, open PDFs, and locate and extract target figures or tables. Retrieved evidence is stored in a memory pool and reused in later turns. The final turn requires multi-evidence fusion across papers, combining visual and textual evidence in the memory pool to produce the structured answer.}
  \label{fig:episode_overview}
\end{figure*}

\textsc{EpiBench} evaluates research agents on episodic multi-turn tasks grounded in multimodal evidence from scientific papers. 
Each episode is a sequence of turns $\{(q_t,\mathcal{I}_t)\}_{t=1}^{T}$, where $q_t$ is an objective question and $\mathcal{I}_t$ optionally provides a seed cue such as an image snippet or a short bibliographic hint. 
The agent answers each $q_t$ and may invoke an external toolkit $\mathcal{T}$ for paper discovery and content access, including searching for papers, opening a paper, and extracting figures, tables or text. 
Figure~\ref{fig:episode_overview} illustrates a representative episode and the corresponding tool-use workflow.

\subsection{Benchmark Construction}
\label{sec:construction}

\begin{figure*}[t]
  \centering
  \begin{minipage}[t]{0.32\textwidth}
    \centering
    \includegraphics[width=\linewidth]{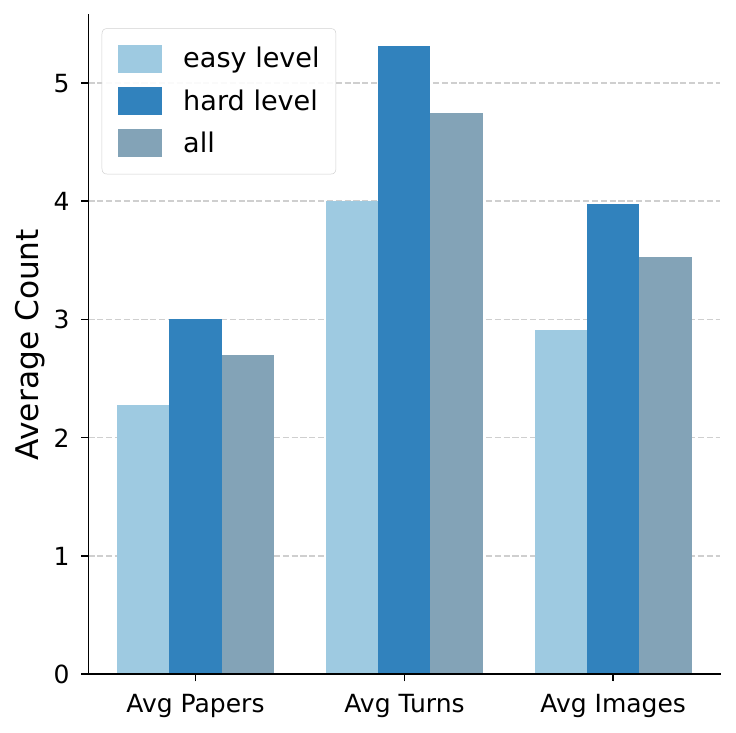}
  \end{minipage}\hfill
  \begin{minipage}[t]{0.32\textwidth}
    \centering
    \includegraphics[width=\linewidth]{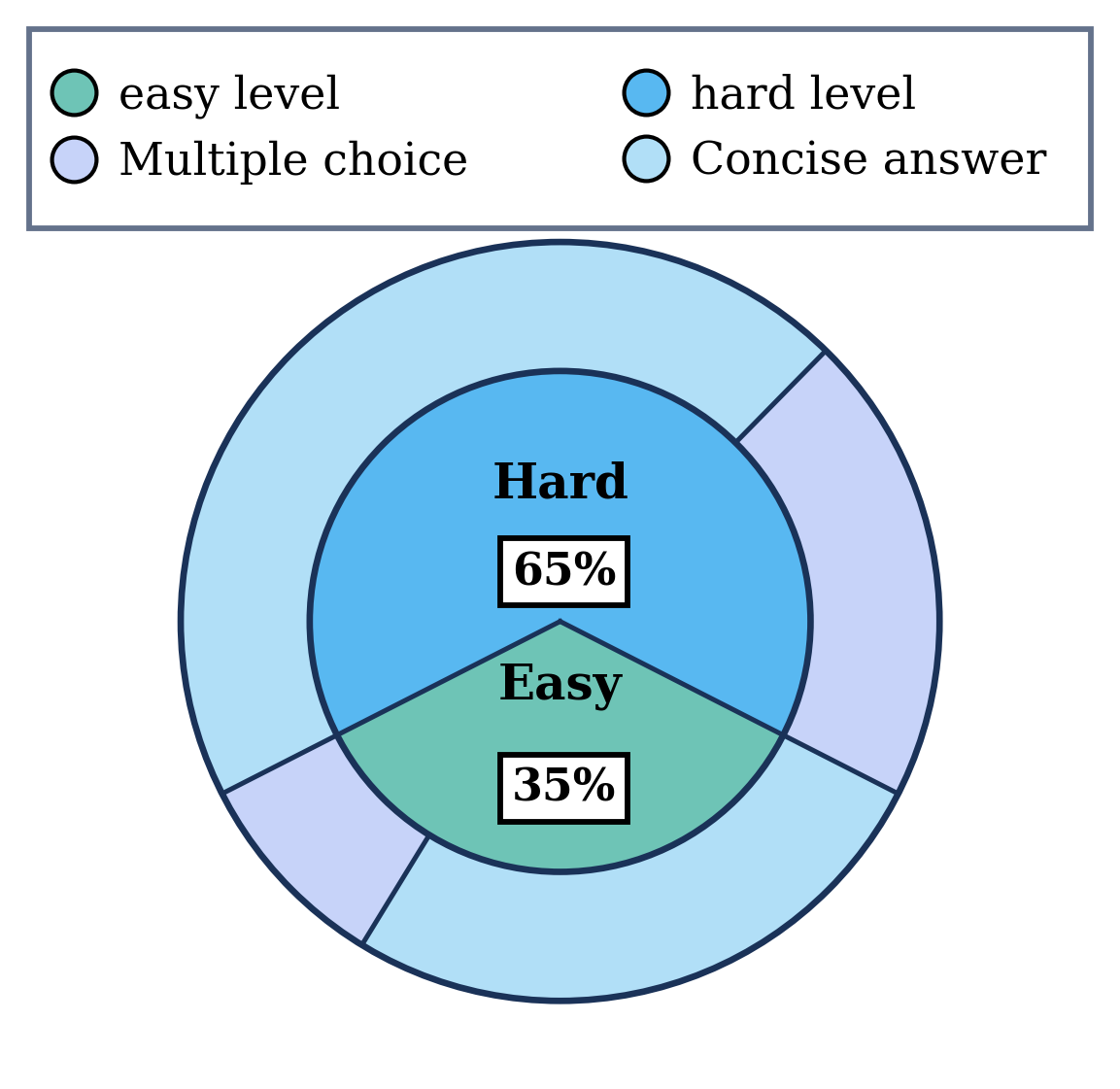}
  \end{minipage}\hfill
  \begin{minipage}[t]{0.32\textwidth}
    \centering
    \includegraphics[width=\linewidth]{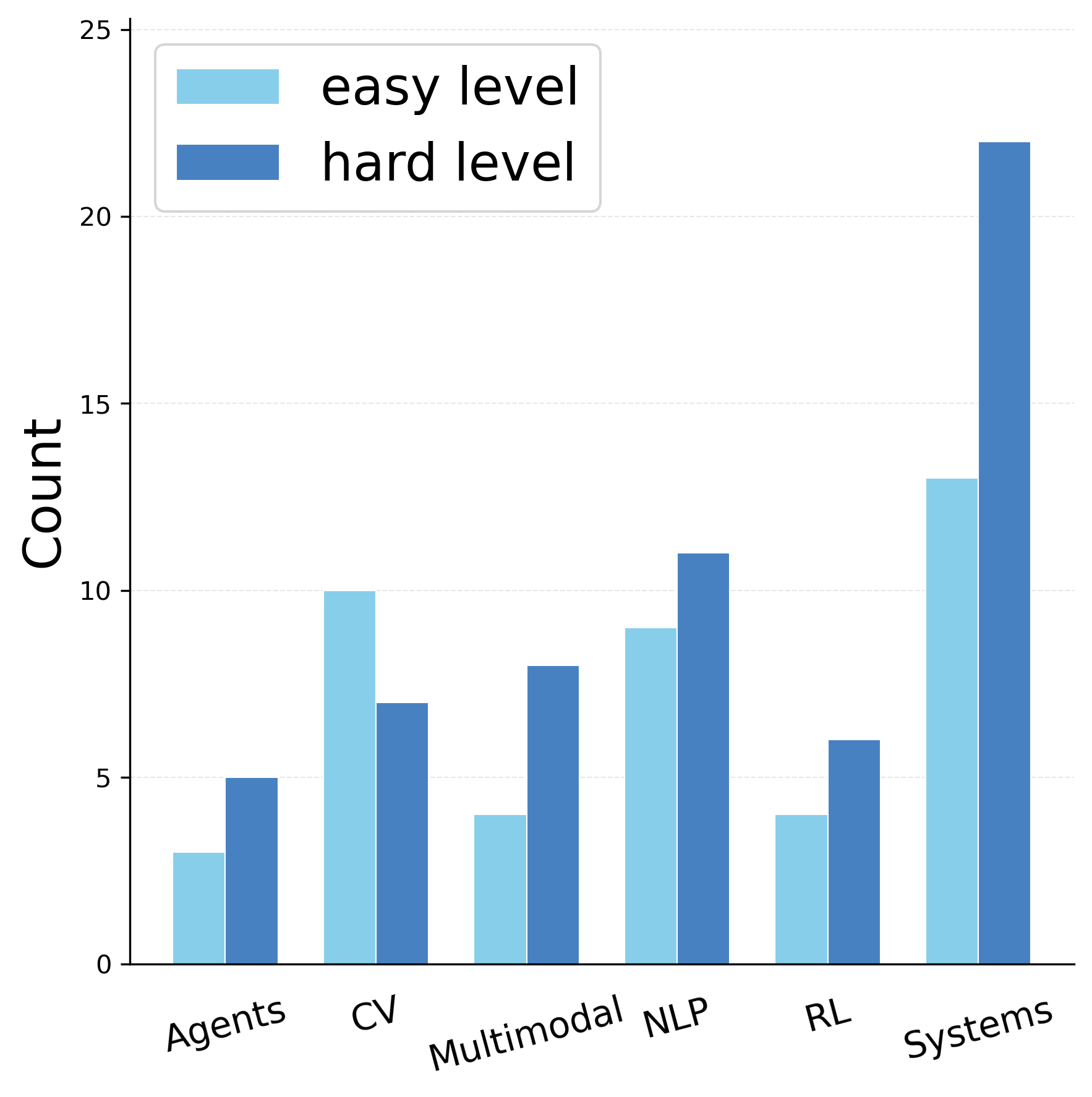}
  \end{minipage}

  \caption{Dataset overview of \textsc{EpiBench}. 
  \textbf{Left:} Statistics of the EpiBench dataset by difficulty level. 
  \textbf{Middle:} Data distribution across categories. 
  \textbf{Right:} Episode counts across subject areas stratified by difficulty.}
  \label{fig:dataset_overview}
\end{figure*}

We construct \textsc{EpiBench} through an expert extraction, expansion, and curation pipeline on papers.
We first collect a seed set of classic papers from public venues and preprint repositories, including OpenReview \cite{openreview} and arXiv \cite{arxiv}.
The seed set covers six broad areas of computer vision and machine learning and contains 68 representative papers.

Next, we expand the candidate pool using Connected Papers \cite{smolyansky2020_connectedpapers}, a citation graph exploration tool that discovers relevant papers around each seed paper at scale.
After deduplication and basic quality filtering, we obtain a pool of 485 papers as the source corpus for episode construction.

We then use GPT-5.2~\cite{openai_gpt5_2_2025} to draft episodic question chains.
Each draft episode is designed to follow a research workflow.
Early turns locate papers and extract evidence from specific figures or tables.
Later turns require reusing previously accessed evidence and integrating multiple evidence units across papers.
For each turn, we also generate a checklist of tool calls, where each call specifies the target paper and, when applicable, the target figure or table.
This supports verification of the intended solution path.

Finally, five annotators with Ph.D.\ degrees in computer science refine, validate, and curate the generated episodes. 
They remove ambiguous cases, correct evidence targets, and resolve annotation inconsistencies through a final consolidation pass, ensuring that each required paper contributes essential evidence to the final answer.
The resulting benchmark contains 102 high-quality episodes, split into Easy (39) and Hard (63).
Difficulty is assigned based on the required number of turns, the amount of cross-paper and multimodal evidence integration, and the extent of evidence reuse required in later turns.
Figure~\ref{fig:dataset_overview} illustrates key statistics of the benchmark.
The left panel shows the average count of papers, turns, and images across different difficulty levels.
The middle panel presents the overall distribution across difficulty levels and answer formats.
The right panel further breaks down the number of episodes by subject area for the easy and hard splits.
Additional details on expert curation and benchmark composition are provided in Appendix~\ref{app:curation} and Appendix~\ref{app:benchmark_composition}.

\subsection{Key Features of \textsc{EpiBench}}
\label{sec:key_features}

\noindent\textbf{Multi-turn research episodes.}
\textsc{EpiBench} evaluates agents on short research workflows rather than isolated questions. 
Each instance is organized as an episode with multiple turns, where later questions depend on evidence acquired earlier. 
This structure captures common research behaviors such as iterative retrieval, progressive refinement of hypotheses, and cross-turn consistency requirements, and it exposes failure modes that do not appear in single-turn settings.

\noindent\textbf{Multimodal evidence integration across papers.}
The benchmark emphasizes evidence fusion from figures, tables, and text. 
Questions are designed so that a correct answer requires aligning information across multiple visual artifacts and, in many cases, across multiple papers linked by citations or indirect cues. 
This setting stresses multimodal grounding, such as reading plotted values and legends, extracting table entries, and reconciling experimental settings across sources.

\noindent\textbf{Proactive citation-guided search.}
\textsc{EpiBench} requires agents to proactively discover relevant papers instead of starting from an explicitly specified target title or identifier. 
Agents must use citation cues and indirect hints accumulated during the episode to formulate search queries, follow reference trails, and resolve paper identities, which better reflects realistic research settings where users rarely provide precise citations.

\noindent\textbf{Evidence reuse with process supervision.}
\textsc{EpiBench} explicitly targets memory use by marking evidence units that should be reused in later turns. 
In particular, for the final high-difficulty turn that requires fusing all previously acquired multimodal evidence, we disable tool access to enforce memory-based reuse rather than repeated re-browsing.
This enables process-level evaluation of whether agents revisit the right evidence and reuse earlier findings correctly when appropriate.

\subsection{Evaluation Metrics}
\label{sec:metrics}

We report four metrics that capture success rate, evidence correctness, minimality gap and error analysis.

\noindent\textbf{Success rate.}
We report episode success rate (ESR), where an episode is counted as correct only if the agent answers all turns correctly. Since the final turn is intentionally harder and requires multi-evidence fusion, we additionally report turn-level accuracy for the final turn (Acc$_{\mathrm{final}}$) and the average accuracy over all non-final turns (Acc$_{\mathrm{pre}}$).

\noindent\textbf{Evidence correctness.}
To verify that correct answers follow the intended reasoning path rather than lucky guesses, we annotate each turn $t$ with a set of required evidence units $\mathcal{E}_{i,t}$, indexed by paper identifier and figure or table identifier, and evaluate whether the agent's accessed evidence $\mathcal{A}_{i,t}$ from the tool trace aligns with $\mathcal{E}_{i,t}$ on correctly answered turns:
\begin{equation}
\mathrm{EC} \;=\;
\frac{\sum_{i=1}^{N}\sum_{t=1}^{T_i} y_{i,t}\,\lvert \mathcal{E}_{i,t}\cap \mathcal{A}_{i,t}\rvert}
{\sum_{i=1}^{N}\sum_{t=1}^{T_i} y_{i,t}\,\lvert \mathcal{E}_{i,t}\rvert}.
\end{equation}
This measures the fraction of required evidence units that are actually accessed when the answer is correct.

\noindent\textbf{Minimality gap.}
Each episode is annotated with a checklist that specifies a minimal viable tool-use path with $c_i^\star$ tool calls.
Let $c_i$ be the number of tool calls executed by the agent in episode $i$, and let $s_i\in\{0,1\}$ indicate episode success.
We report the minimality gap on successful episodes:
\begin{equation}
\mathrm{MG}_{\mathrm{succ}} \;=\; \frac{1}{\sum_{i=1}^{N} s_i}\sum_{i=1}^{N} s_i \cdot \frac{c_i}{c_i^\star}.
\end{equation}
A larger gap indicates more redundant tool usage relative to the minimal viable solution path.

\noindent\textbf{Error analysis.}
To understand failure modes beyond aggregate accuracy, we perform trace-based error attribution. For each failed episode, we assign a single error label based on the earliest point of failure in the tool trace and evidence alignment, and then report the proportion of episodes in each category.
Representative failure cases are provided in Appendix~\ref{app:failure_cases}.
\section{Agent Framework}
\label{sec:blind}

This section describes the agent implementation used in our experiments. Our goal is to provide a clean and reproducible interface that couples an LMM with external tools and a persistent memory, and supports detailed logging for process-level evaluation. We build on \textsc{smolagents}~\cite{smolagents} and adapt it to our episodic paper-centric setting.

\subsection{Agentic Workflow}
\label{sec:agent_workflow}

At each turn $t$, the agent receives a question $q_t$ and optional images. It maintains a memory state $M_t$ that stores all previously observed content in the episode. The agent follows an iterative loop that alternates between tool invocation and answering. 
In practice, we implement this loop using function-calling. The model decides whether to call a tool and how to parameterize the call based on the current question and memory context. 

\subsection{Tool Environment}
\label{sec:tool_env}

We provide four tools tailored to literature search and paper content access.

\begin{itemize}
  \item \textbf{\textsc{Web Search}.} Queries a search backend and returns ranked results with titles, snippets, and URLs. We use both Google search and DuckDuckGo search as backends~\cite{google_custom_search_json_api,duckduckgo_search_2026}. This tool is used to locate candidate papers, resolve paper identities from indirect cues, and follow citation leads.

  \item \textbf{\textsc{PDF Extractor}}~\cite{wang2024mineruopensourcesolutionprecise}. Given a paper identifier (for example, title or arXiv ID), this tool downloads the PDF and parses it into a normalized markdown representation.

  \item \textbf{\textsc{PDF Extractor RAG}.} Performs retrieval over the cached markdown for a given paper. Given a text query, it returns the most relevant spans together with lightweight provenance, such as section context and figure or table references. This tool supports targeted evidence lookup without re-reading the full paper.

  \item \textbf{\textsc{Content Extractor}.} Extracts a specified figure or table from a cached paper. For figures, it returns the figure image and caption context. For tables, it returns a structured table representation together with surrounding text when available. This tool is used when a question requires visual grounding.
\end{itemize}

All tools share a unified interface and return structured outputs that can be appended to memory and traced during evaluation. A more detailed tool specification is provided in Appendix~\ref{app:tool_spec}.

\subsection{Memory Management}
\label{sec:memory_mgmt}

Memory is implemented as an episode-level chronological store. We append retrieved PDFs, extracted figures and tables, and tool observations in the order they are accessed. This design keeps the agent state transparent and enables post hoc computation of reuse and efficiency metrics from the interaction trace.

To stress evidence reuse, we run the final turn in a memory-only mode for episodes that require reuse. In this mode, we disable all tools. The agent must answer using evidence already stored in memory rather than re-browsing or re-downloading documents. This setting isolates memory utilization from additional retrieval and makes reuse behavior measurable and comparable across models.

\section{Experiment}

\subsection{Experimental Setup}
\label{sec:exp_setup}

\noindent\textbf{Models.}
We evaluate representative four open-source and four closed-source LMMs as agent backbones.
Our open-source baselines include Qwen3-VL-235B-A22B-Instruct, Qwen3-VL-235B-A22B-Thinking~\cite{bai2025qwen3vltechnicalreport}, GLM-4.5V~\cite{vteam2026glm45vglm41vthinkingversatilemultimodal}, and Kimi-K2.5~\cite{kimiteam2026kimik25visualagentic}.
Our closed-source baselines include GPT-5-Mini~\cite{singh2025openaigpt5card}, GPT-5.2~\cite{openai_gpt5_2_2025}, Grok-4.1~\cite{xai_grok41_2025}, and Gemini-2.5-Pro~\cite{comanici2025gemini25pushingfrontier}.
We also include a human baseline from two computer-science Ph.D. experts, who solve the same episodes under the same tool interface.

\noindent\textbf{Evaluation Metrics.}
We report three primary outcome metrics in the main table, episode success rate (ESR), accuracy on the final turn (Acc$_{\mathrm{final}}$), and accuracy on all non-final turns (Acc$_{\mathrm{pre}}$). Full metric definitions are given in Sec.~\ref{sec:metrics}.
All turn answers are scored with the LLM-as-a-Judge~\cite{liu2025amelostableframeworkarenabased}, where GPT-5.2~\cite{openai_gpt5_2_2025} assigns binary correctness labels under a fixed rubric.
In addition, we present diagnostic analyses using evidence correctness and minimality gap, together with an error analysis that attributes failures to retrieval, perception, memory reading, or reasoning.
Further details on judge agreement and evaluation reliability are reported in Appendix~\ref{app:evaluation_reliability}.

\noindent\textbf{Implementation Details.}
We employ a ReAct-like\cite{yao2023react} workflow, in which the agent first generates a plan outlining its intended actions before producing executable code for tool invocation. For each turn within an episode, the agent is constrained by a limited step budget (10 steps by default). If the agent fails to provide a final answer within the allotted steps, it is granted one additional opportunity to summarize its exploration process and deliver an alternative answer in a single response.
For generation, we use the default decoding configuration for all evaluated models and enable their reasoning mode when available. The full evaluation prompt template is included in Appendix~\ref{app:evaluation_prompt}.

\subsection{Main Results}
\label{sec:main_results}

Table~\ref{tab:main_results} summarizes the main results on \textsc{EpiBench} across eight LMMs under our agentic workflow. Across all models, GPT-5.2 achieves the best overall performance, and it is also the strongest model on the Hard split in terms of ESR, reaching 29.23\%. 
However, even for the most capable closed-source systems, high turn-level accuracy on non-final turns does not translate into reliable end-to-end performance. 
Acc$_{\mathrm{pre}}$ remains relatively high for several closed-source models, yet Acc$_{\mathrm{final}}$ and ESR drop sharply, indicating that multi-evidence fusion across papers and figures or tables, together with memory-based evidence reuse, remains a major bottleneck. Aggregate runtime and token-cost statistics are reported in Appendix~\ref{app:efficiency}.

\begin{table}[t]
\caption{Main results on \textsc{EpiBench}. We report episode success rate (ESR, \%), accuracy on the final turn (Acc$_{\mathrm{final}}$, \%), and accuracy on all non-final turns (Acc$_{\mathrm{pre}}$, \%), stratified by difficulty. Instr. denotes instruction-tuned models and Think. denotes thinking-mode models.}
\label{tab:main_results}
\centering
\scriptsize
\setlength{\tabcolsep}{2.5pt}   
\renewcommand{\arraystretch}{1.3}
\begin{tabular}{lccccccccc}
\toprule
\multirow{2}{*}{Base LMM} &
\multicolumn{3}{c}{Easy} &
\multicolumn{3}{c}{Hard} &
\multicolumn{3}{c}{Average} \\
\cmidrule(lr){2-4}\cmidrule(lr){5-7}\cmidrule(lr){8-10}
& ESR & Acc$_{\mathrm{final}}$ & Acc$_{\mathrm{pre}}$
& ESR & Acc$_{\mathrm{final}}$ & Acc$_{\mathrm{pre}}$
& ESR & Acc$_{\mathrm{final}}$ & Acc$_{\mathrm{pre}}$ \\
\midrule

\multicolumn{10}{c}{\textit{Open-source models}} \\
\midrule
Qwen3-VL-235B-Instr.  & 16.98 & 41.51 & 70.67 & 8.16 & \textbf{51.02} & 49.77 & 12.75 & 46.08 & 55.17 \\
Qwen3-VL-235B-Think.  & 30.00 & 65.00 & 76.09 & 3.23 & 43.55 & 43.84 & 13.73 & 51.96 & 51.90 \\
GLM-4.5V              & 8.11  & 51.35 & 38.95 & 0.00 & 30.77 & 29.17 & 2.94  & 38.24 & 31.59 \\
Kimi-K2.5      & 52.94 & 73.53 & 89.77 & 26.15 & 49.23 & 79.51 & 35.35 & 57.58 & 81.91 \\
\midrule

\multicolumn{10}{c}{\textit{Closed-source models}} \\
\midrule
GPT-5-Mini            & 29.73 & 62.16 & 84.21 & 21.54 & 50.77 & 74.65 & 24.51 & 54.90 & 77.02 \\
GPT-5.2        & \textbf{58.33} & \textbf{77.78} & \textbf{91.30} & \textbf{29.23} & 49.23 & \textbf{86.46} & \textbf{39.60} & \textbf{59.41} & \textbf{87.63} \\
Grok-4.1        & 15.38    & 66.67    & 63.10    & 6.35    & 39.68    & 51.07    & 9.80    & 50.00    & 53.85    \\
Gemini-2.5-Pro         & 27.03 & 64.86 & 75.79 & 7.69 & 46.15 & 57.29 & 14.71 & 52.94 & 61.88 \\
\midrule

\multicolumn{10}{c}{\textit{Human Baseline}} \\
\midrule
\rowcolor{gray!15}
Ph.D.\ Experts & 91.43 & 97.14 & 96.67 & 81.36 & 94.42 & 95.79 & 85.11 & 95.74 & 96.01 \\
\bottomrule\end{tabular}
\end{table}

Notably, Acc$_{\mathrm{final}}$ is particularly informative in our setting because the final turn is designed to require reuse of previously retrieved evidence and joint reasoning over multiple evidence units. The persistent gap between Acc$_{\mathrm{pre}}$ and Acc$_{\mathrm{final}}$, together with the low ESR on hard episodes, suggests that failures often arise after initial evidence acquisition, when agents must organize, recall, and align evidence under constraints. In contrast, two Ph.D.\ experts achieve substantially higher performance, highlighting a large gap between current agents and expert research workflows. To understand where models break down, we next present a process-level analysis of tool use and evidence chains in Sec.~\ref{sec:analysis}, followed by a focused study of memory reuse behavior and its impact on success in Sec.~\ref{sec:analysis_memory}.

\begin{table}[t]
\caption{Process-level diagnostics on \textsc{EpiBench}. We report evidence correctness (EC, \%) and minimality gap (MG) by difficulty. Higher EC is better. Lower MG is better. Instr. denotes instruction-tuned models and Think. denotes thinking-mode models.}
\label{tab:ec_mg}
\centering
\scriptsize
\renewcommand{\arraystretch}{1.3}

\begin{tabular*}{\linewidth}{@{\extracolsep{\fill}} lcccccc}
\toprule
\textbf{Base LMM} &
\multicolumn{2}{c}{Easy} &
\multicolumn{2}{c}{Hard} &
\multicolumn{2}{c}{Average} \\
\cmidrule(lr){2-3}\cmidrule(lr){4-5}\cmidrule(lr){6-7}
& EC $\uparrow$ & MG $\downarrow$
& EC $\uparrow$ & MG $\downarrow$
& EC $\uparrow$ & MG $\downarrow$ \\
\midrule

\multicolumn{7}{c}{\textit{Open-source models}} \\
\midrule
Qwen3-VL-235B-Instr.  & 64.63 & 1.41 & 63.48 & 1.03 & 63.85 & 1.23 \\
Qwen3-VL-235B-Think.  & 54.55 & \textbf{1.02} & 45.66 & \textbf{1.00} & 48.63 & \textbf{1.01} \\
GLM-4.5V              & 24.49 & 2.60 & 31.20 & 2.60 & 29.31 & 2.60 \\
Kimi-K2.5      & 81.30 & 1.49 & 84.91 & 1.20 & 84.13 & 1.35 \\
\midrule

\multicolumn{7}{c}{\textit{Closed-source models}} \\
\midrule
GPT-5-Mini            & 75.40 & 3.00 & 82.43 & 1.94 & 80.75 & 2.32 \\
GPT-5.2        & \textbf{85.93} & 1.34 & \textbf{86.15} & 1.27 & \textbf{86.10} & 1.30 \\
Grok-4.1         & 85.71    & 3.33   & 78.33    & 2.88   & 80.12    & 3.07   \\
Gemini-2.5-Pro         & 71.30 & 2.17 & 68.71 & 1.41 & 69.39 & 1.80 \\
\midrule

\multicolumn{7}{c}{\textit{Human Baseline}} \\
\midrule
Ph.D. Experts          & 83.45 & 1.00 & 88.35 & 1.02 & 86.88 & 1.01 \\
\bottomrule
\end{tabular*}
\end{table}

\subsection{Process-Level Diagnostics and Error Analysis}
\label{sec:analysis}

To understand why high turn-level accuracy does not translate into high episode success, we analyze process-level behaviors beyond final answers. Table~\ref{tab:ec_mg} reports evidence correctness (EC), measuring alignment to required evidence units, and minimality gap (MG), measuring tool-call redundancy relative to the minimal viable checklist. GPT-5.2 and Kimi-K2.5 achieve consistently high EC, suggesting that their correct answers are more often supported by sufficient evidence, whereas other models more frequently rely on prior knowledge or guessing. MG varies widely and is weakly coupled with EC, and redundancy is often higher on Easy than on Hard, consistent with over-exploration on simpler tasks. Compared with Ph.D.\ experts, both EC and MG reveal a substantial gap, motivating finer-grained error analysis from interaction traces.

Notably, EC and MG quantify whether agents follow the intended evidence path and \emph{how} efficiently they use tools, but they do not explain \emph{why} episodes fail. We therefore conduct an error-rate analysis by attributing each failed episode to its earliest point of deviation in the tool trace and evidence alignment. Specifically, based on their underlying causes, we categorize the failures into five types:
\begin{wrapfigure}{r}{0.48\linewidth}
  \centering
  \includegraphics[width=\linewidth]{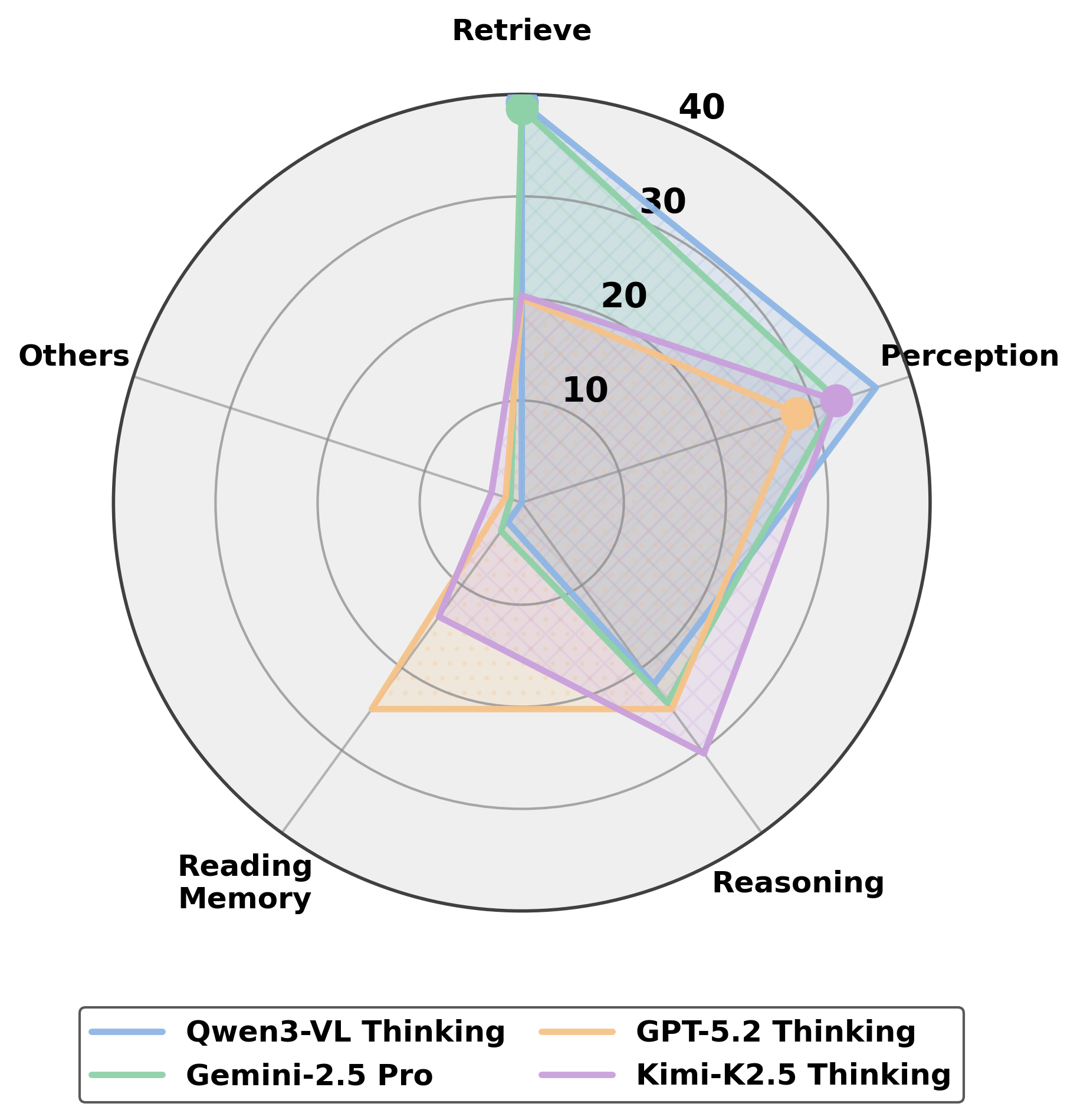}
  \caption{Failure-type distribution (\%) across models.}
  \label{fig:error_radar}
\end{wrapfigure}
\textbf{Retrieve} errors arise when the agent fails to locate the intended papers.
\textbf{Perception} errors occur when the agent accesses the relevant evidence but misreads or misinterprets figures or tables. \textbf{Reasoning} errors refer to incorrect integration or constraint satisfaction despite having accessed the necessary evidence. \textbf{Reading Memory} errors capture cases where the required evidence was retrieved earlier but is not correctly reused at later turns, including conflation across sources or missing evidence selection during final fusion. 
and \textbf{Others} collects residual failures that do not fit the above categories, such as max-step terminations, runtime errors or pdf-parsing errors.

Figure~\ref{fig:error_radar} summarizes failure-type distributions across models. We observe a clear shift in where different agents break down. Gemini-2.5-Pro and Qwen3-VL-Thinking are dominated by \textbf{Retrieve} errors, suggesting that they often fail before reaching the stage where multi-evidence fusion becomes decisive. In contrast, stronger reasoning-oriented models shift mass toward later-stage failures. GPT-5.2 and Kimi-K2.5 show larger shares of \textbf{Perception} and \textbf{Reasoning} errors, indicating that once retrieval succeeds, the remaining bottleneck lies in reliable multimodal evidence alignment and integration. GPT-5.2 also exhibits a higher \textbf{Reading Memory} error rate, which is consistent with it reaching memory-reuse turns more often due to stronger early evidence acquisition. This still highlights substantial room for improvement in multimodal evidence reuse from memory.

\begin{figure}[t]
  \centering
  \begin{minipage}[t]{0.49\linewidth}
    \centering
    \includegraphics[width=\linewidth]{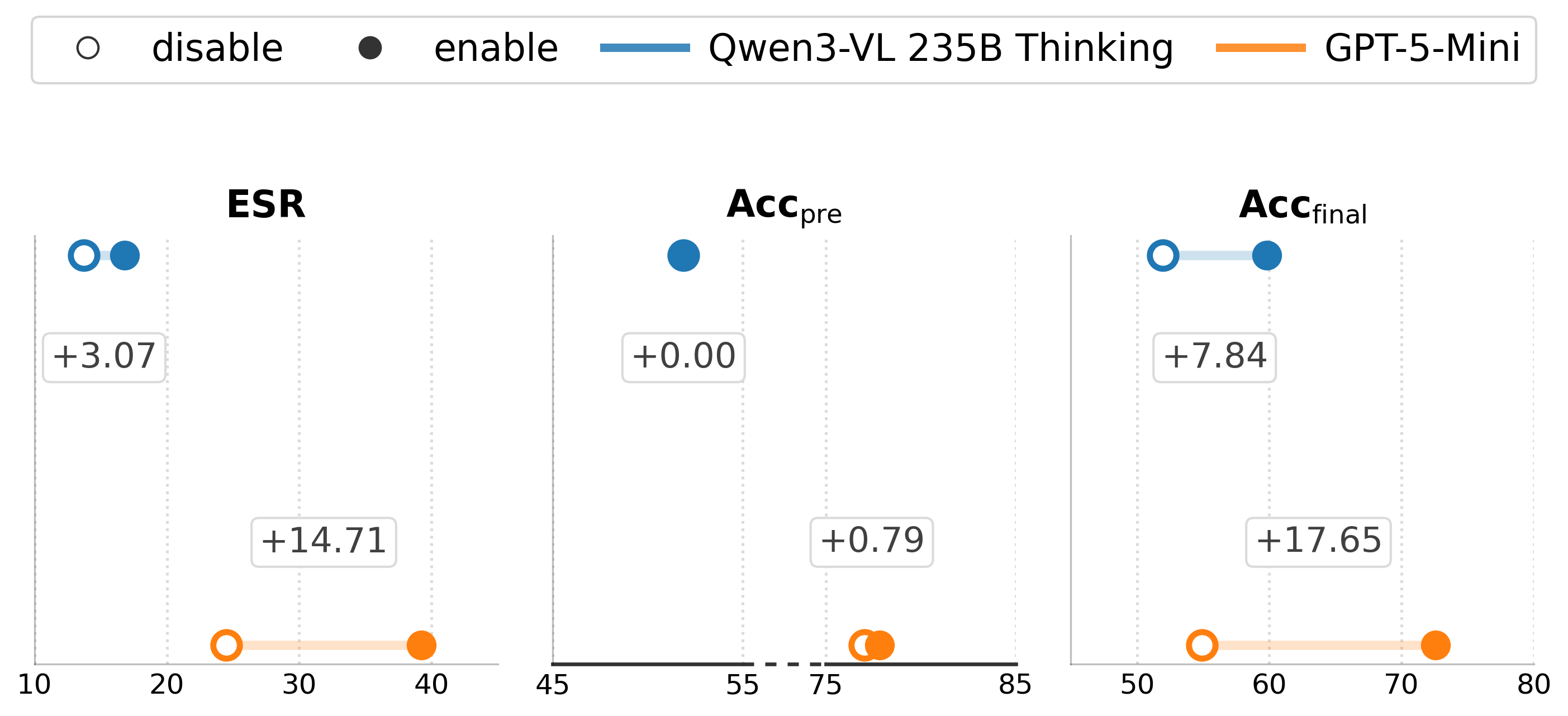}
    \caption*{\textbf{(a)} Final-turn tool access.}
  \end{minipage}\hfill
  \begin{minipage}[t]{0.49\linewidth}
    \centering
    \includegraphics[width=\linewidth]{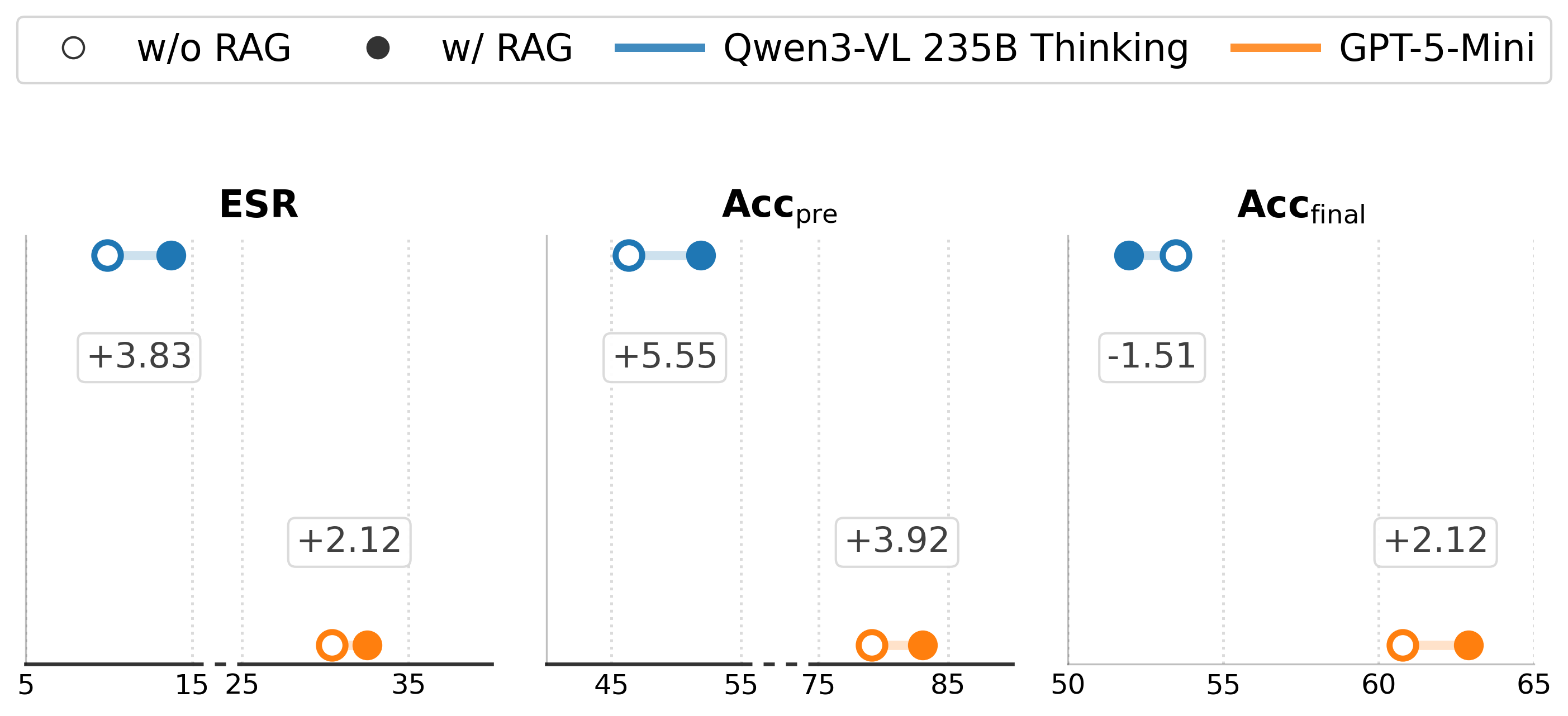}
    \caption*{\textbf{(b)} Removing PDF text RAG.}
  \end{minipage}
  \caption{Ablations on memory-centric constraints. \textbf{(a)} Allowing tool use in the final turn yields a large recovery in ESR and Acc$_{\mathrm{final}}$, indicating that multimodal evidence cached across turns is often insufficient for reliable final fusion under the memory-only protocol. \textbf{(b)} Removing PDF text RAG has mixed effects and generally smaller impact, suggesting that the primary bottleneck lies in multimodal evidence selection and alignment rather than text retrieval alone.}
  \label{fig:memory_ablations}
\end{figure}

\subsection{Impact of Evidence Memory and Final-Turn Constraints}
\label{sec:analysis_memory}

We probe how memory constraints affect episodic success with two ablations in settings where \textsc{EpiBench} requires multi-evidence fusion across turns. First, we compare the main \emph{memory-only} protocol (tools disabled on the final turn) with a relaxed setting that \emph{enables} tool use on the final turn, which serves as an upper-bound style reference when agents are allowed to re-open and re-browse documents. Second, we remove the PDF text RAG tool to isolate the contribution of within-document text retrieval.

As shown in Fig.~\ref{fig:memory_ablations}a, allowing tool use on the final turn substantially improves ESR and Acc$_{\mathrm{final}}$, suggesting that memory-based reuse and multi-evidence integration remain a primary bottleneck.
Figure~\ref{fig:memory_ablations}b shows that removing text RAG has smaller and model-dependent effects, suggesting that text retrieval is not the main limiting factor for these episodes. Overall, the results point to multimodal evidence selection, indexing, and alignment in memory as the primary challenge for reliable research-workflow completion.

\subsection{Effect of Reasoning Step Budget}
\label{sec:analysis_steps}

\begin{figure}[t]
  \centering
  \begin{minipage}[t]{0.49\linewidth}
    \centering
    \includegraphics[width=\linewidth]{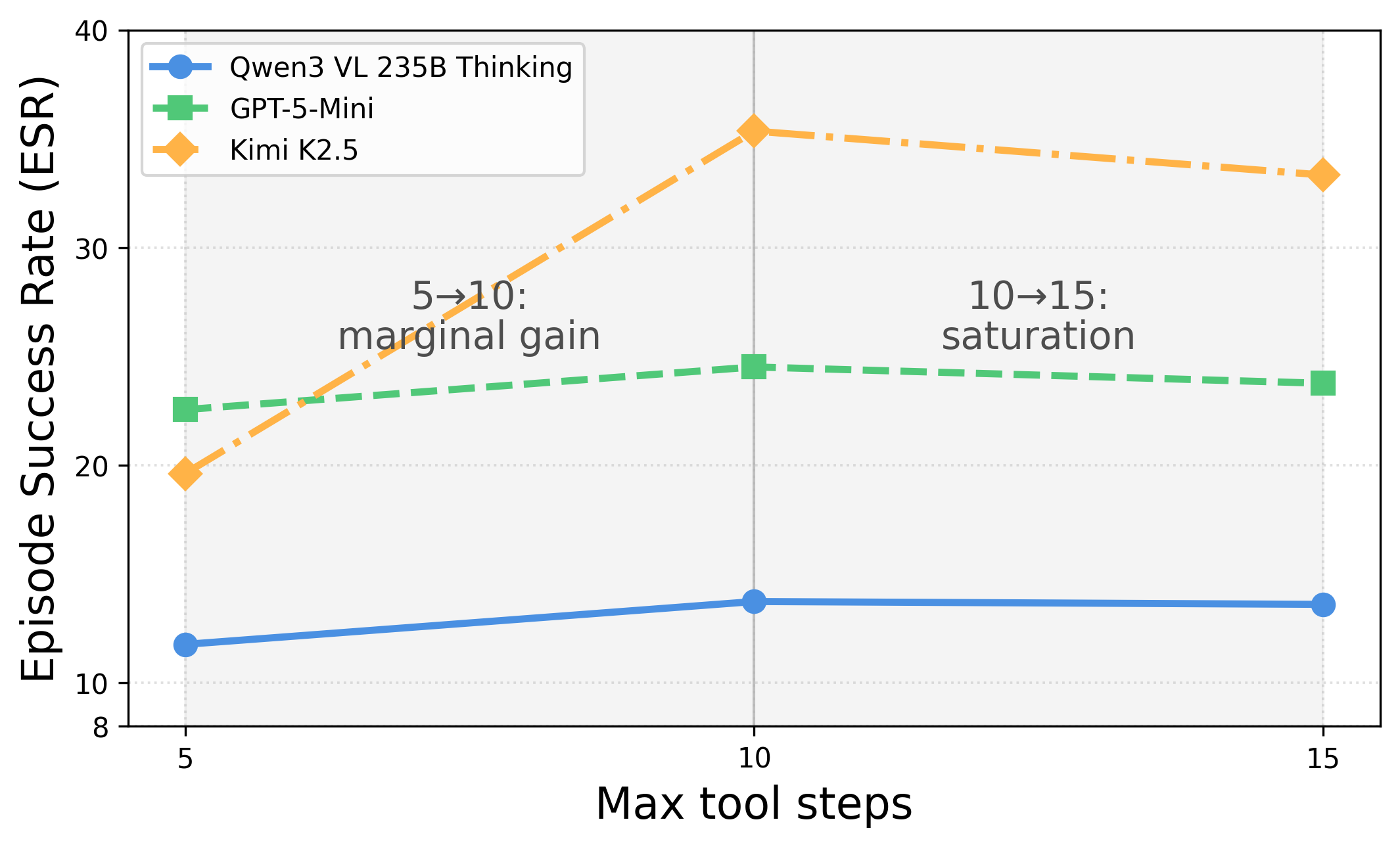}
    \caption*{\textbf{(a)} Episode success rate (ESR).}
  \end{minipage}\hfill
  \begin{minipage}[t]{0.49\linewidth}
    \centering
    \includegraphics[width=\linewidth]{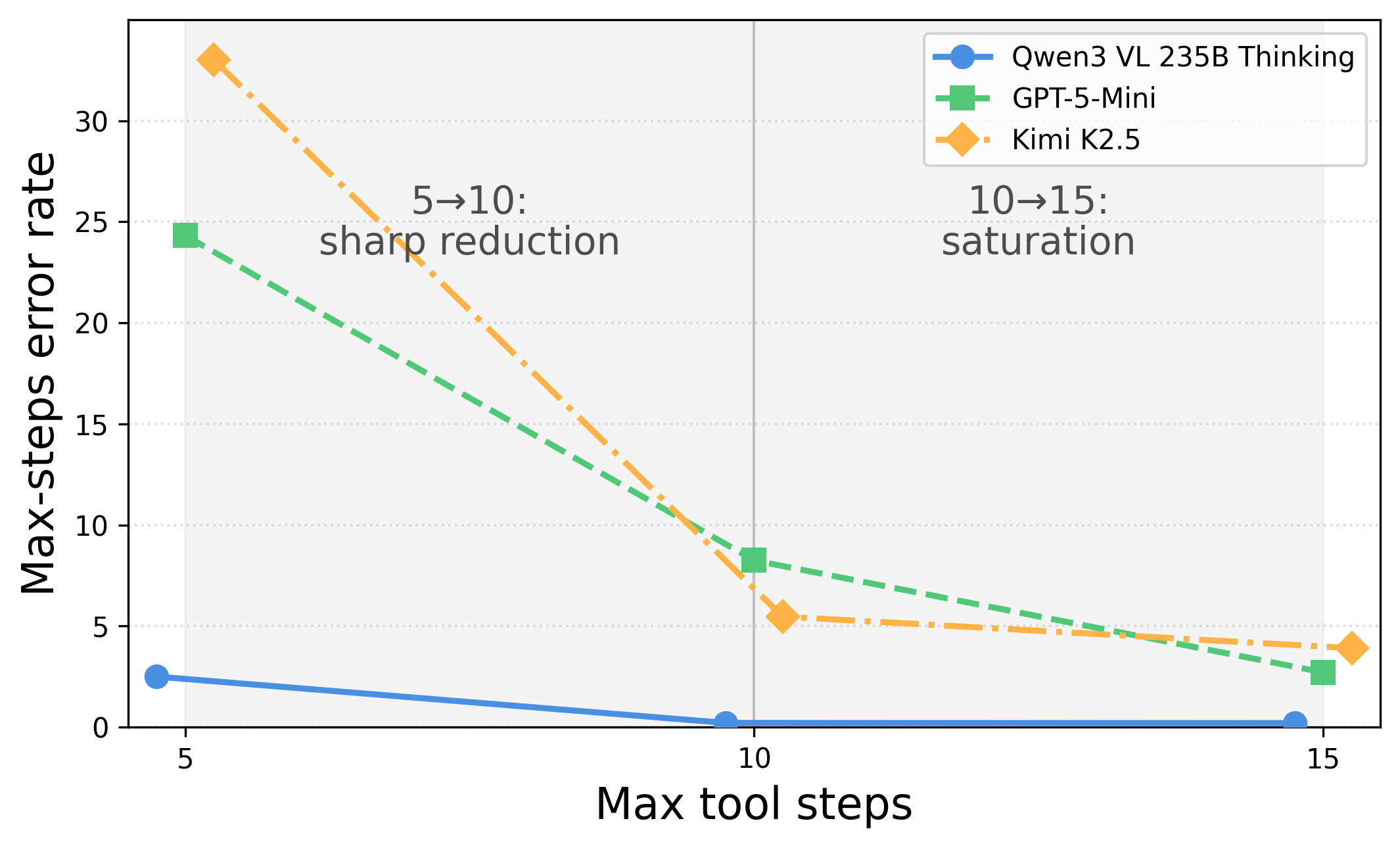}
    \caption*{\textbf{(b)} Max-steps error rate.}
  \end{minipage}
  \caption{Effect of step budget $S_{\max}$ on episodic performance. \textbf{(a)} ESR increases from 5 to 10 steps but saturates from 10 to 15. \textbf{(b)} Max-steps error rate drops sharply from 5 to 10, indicating fewer budget-induced terminations, while additional budget yields diminishing returns.}
  \label{fig:step_budget}
\end{figure}

We analyze how the step budget affects episodic performance by varying the maximum number of tool calls $S_{\max}\in\{5,10,15\}$ while keeping all other settings fixed. Figure~\ref{fig:step_budget} reports episode success rate (ESR) together with the max-steps error rate, which captures failures caused by exceeding the step limit.

Figure~\ref{fig:step_budget}a shows a consistent improvement in ESR when increasing the budget from 5 to 10 across the evaluated models, but little to no additional gain when further increasing to 15, indicating a saturation regime. Figure~\ref{fig:step_budget}b explains part of this behavior: max-steps errors drop sharply from 5 to 10, suggesting that a tighter budget often truncates episodes before agents complete evidence acquisition and recovery steps. In contrast, the decrease from 10 to 15 is small while ESR remains nearly unchanged, implying that remaining failures are not primarily due to insufficient steps. Instead, beyond a moderate budget, performance is limited by systematic issues in evidence grounding, cross-evidence alignment, and memory-based evidence reuse that additional tool calls do not resolve.

\section{Conclusion}
\label{sec:conclusion}

We introduced \textsc{EpiBench}, an episodic multi-turn multimodal benchmark that evaluates research agents in short, process-level workflows requiring proactive cross-paper navigation, visual evidence extraction, and memory-based reuse for objective multi-evidence questions. 
By emphasizing cross-paper multi-figure and multi-table integration under workflow constraints, \textsc{EpiBench} provides fine-grained evaluation and diagnostics beyond single-turn accuracy. 
Experiments on leading open-source and closed-source agents show that current systems remain far from reliable research assistance, with the best model reaching only 29.23\% on the hard split, highlighting persistent bottlenecks in evidence grounding, multi-evidence fusion, and robust reuse of previously acquired evidence. 
We hope \textsc{EpiBench} will serve as a practical platform for developing more verifiable, efficient, and reproducible research agents. Additional reproducibility details are provided in Appendix~\ref{app:reproducibility}.

%
%
\bibliographystyle{splncs04}
\bibliography{main}

@misc{liu2025amelostableframeworkarenabased,
      title={am-ELO: A Stable Framework for Arena-based LLM Evaluation}, 
      author={Zirui Liu and Jiatong Li and Yan Zhuang and Qi Liu and Shuanghong Shen and Jie Ouyang and Mingyue Cheng and Shijin Wang},
      year={2025},
      eprint={2505.03475},
      archivePrefix={arXiv},
      primaryClass={cs.AI},
      url={https://arxiv.org/abs/2505.03475}, 
}

@misc{openreview,
  title        = {OpenReview},
  howpublished = {\url{https://openreview.net/}},
  note         = {Accessed: 2026-02-27}
}

@misc{arxiv,
  title        = {arXiv},
  howpublished = {\url{https://arxiv.org/}},
  note         = {Accessed: 2026-02-27}
}

@misc{smolyansky2020_connectedpapers,
  author       = {Smolyansky, Eddie},
  title        = {Announcing Connected Papers — a visual tool for researchers to find and explore academic papers},
  year         = {2020},
  month        = jun,
  day          = {2},
  howpublished = {\url{https://medium.com/connectedpapers/announcing-connected-papers-a-visual-tool-for-researchers-to-find-and-explore-academic-papers-89146a54c7d4}},
  note         = {Accessed: 2026-02-27}
}

@misc{google_custom_search_json_api,
  title        = {Custom Search JSON API: Use REST to Invoke the API},
  author       = {{Google for Developers}},
  howpublished = {\url{https://developers.google.com/custom-search/v1/using_rest}},
  note         = {Accessed 2026-03-02}
}

@misc{vishesh2025aegisagentextractiongeographic,
      title={AEGIS: An Agent for Extraction and Geographic Identification in Scholarly Proceedings}, 
      author={Om Vishesh and Harshad Khadilkar and Deepak Akkil},
      year={2025},
      eprint={2509.09470},
      archivePrefix={arXiv},
      primaryClass={cs.LG},
      url={https://arxiv.org/abs/2509.09470}, 
}

@inproceedings{10.1145/3747912.3747962,
author = {Mikriukov, Andrei and Senokosov, Artsiom and Succi, Giancarlo and Tormasov, Alexander and Plaksin, Yaroslav and Trofimova, Ekaterina and Sitnikov, Vladimir},
title = {AI Tools for Automating Systematic Literature Reviews},
year = {2025},
isbn = {9798400715136},
publisher = {Association for Computing Machinery},
address = {New York, NY, USA},
url = {https://doi.org/10.1145/3747912.3747962},
doi = {10.1145/3747912.3747962},
booktitle = {Proceedings of the 2025 International Conference on Software Engineering and Computer Applications},
pages = {25–30},
numpages = {6},
location = {
},
series = {SECA '25}
}

@misc{wang2026paperarenaevaluationbenchmarktoolaugmented,
      title={PaperArena: An Evaluation Benchmark for Tool-Augmented Agentic Reasoning on Scientific Literature}, 
      author={Daoyu Wang and Mingyue Cheng and Shuo Yu and Zirui Liu and Ze Guo and Xin Li and Qi Liu},
      year={2026},
      eprint={2510.10909},
      archivePrefix={arXiv},
      primaryClass={cs.AI},
      url={https://arxiv.org/abs/2510.10909}, 
}

@misc{huang2026mmdeepresearchbenchbenchmarkmultimodaldeep,
      title={MMDeepResearch-Bench: A Benchmark for Multimodal Deep Research Agents}, 
      author={Peizhou Huang and Zixuan Zhong and Zhongwei Wan and Donghao Zhou and Samiul Alam and Xin Wang and Zexin Li and Zhihao Dou and Li Zhu and Jing Xiong and Chaofan Tao and Yan Xu and Dimitrios Dimitriadis and Tuo Zhang and Mi Zhang},
      year={2026},
      eprint={2601.12346},
      archivePrefix={arXiv},
      primaryClass={cs.CV},
      url={https://arxiv.org/abs/2601.12346}, 
}

@inproceedings{
yao2023react,
title={ReAct: Synergizing Reasoning and Acting in Language Models},
author={Shunyu Yao and Jeffrey Zhao and Dian Yu and Nan Du and Izhak Shafran and Karthik R Narasimhan and Yuan Cao},
booktitle={The Eleventh International Conference on Learning Representations },
year={2023},
url={https://openreview.net/forum?id=WE_vluYUL-X}
}

@inproceedings{NEURIPS2023_d842425e,
 author = {Schick, Timo and Dwivedi-Yu, Jane and Dessi, Roberto and Raileanu, Roberta and Lomeli, Maria and Hambro, Eric and Zettlemoyer, Luke and Cancedda, Nicola and Scialom, Thomas},
 booktitle = {Advances in Neural Information Processing Systems},
 editor = {A. Oh and T. Naumann and A. Globerson and K. Saenko and M. Hardt and S. Levine},
 pages = {68539--68551},
 publisher = {Curran Associates, Inc.},
 title = {Toolformer: Language Models Can Teach Themselves to Use Tools},
 url = {https://proceedings.neurips.cc/paper_files/paper/2023/file/d842425e4bf79ba039352da0f658a906-Paper-Conference.pdf},
 volume = {36},
 year = {2023}
}

@misc{qin2023toolllmfacilitatinglargelanguage,
      title={ToolLLM: Facilitating Large Language Models to Master 16000+ Real-world APIs}, 
      author={Yujia Qin and Shihao Liang and Yining Ye and Kunlun Zhu and Lan Yan and Yaxi Lu and Yankai Lin and Xin Cong and Xiangru Tang and Bill Qian and Sihan Zhao and Lauren Hong and Runchu Tian and Ruobing Xie and Jie Zhou and Mark Gerstein and Dahai Li and Zhiyuan Liu and Maosong Sun},
      year={2023},
      eprint={2307.16789},
      archivePrefix={arXiv},
      primaryClass={cs.AI},
      url={https://arxiv.org/abs/2307.16789}, 
}

@Misc{smolagents,
  title =        {`smolagents`: a smol library to build great agentic systems.},
  author =       {Aymeric Roucher and Albert Villanova del Moral and Thomas Wolf and Leandro von Werra and Erik Kaunismäki},
  howpublished = {\url{https://github.com/huggingface/smolagents}},
  year =         {2025}
}

@misc{duckduckgo_search_2026,
  author       = {{DuckDuckGo}},
  title        = {DuckDuckGo Search Engine},
  howpublished = {\url{https://duckduckgo.com/}},
  year         = {2026},
  note         = {Accessed: 2026-03-03}
}

@misc{yan2025mmcradvancingvisuallanguage,
      title={MMCR: Advancing Visual Language Model in Multimodal Multi-Turn Contextual Reasoning}, 
      author={Dawei Yan and Yang Li and Qing-Guo Chen and Weihua Luo and Peng Wang and Haokui Zhang and Chunhua Shen},
      year={2025},
      eprint={2503.18533},
      archivePrefix={arXiv},
      primaryClass={cs.AI},
      url={https://arxiv.org/abs/2503.18533}, 
}

@misc{zeng2026visiondeepresearchbenchmarkrethinkingvisual,
      title={Vision-DeepResearch Benchmark: Rethinking Visual and Textual Search for Multimodal Large Language Models}, 
      author={Yu Zeng and Wenxuan Huang and Zhen Fang and Shuang Chen and Yufan Shen and Yishuo Cai and Xiaoman Wang and Zhenfei Yin and Lin Chen and Zehui Chen and Shiting Huang and Yiming Zhao and Xu Tang and Yao Hu and Philip Torr and Wanli Ouyang and Shaosheng Cao},
      year={2026},
      eprint={2602.02185},
      archivePrefix={arXiv},
      primaryClass={cs.CV},
      url={https://arxiv.org/abs/2602.02185}, 
}

@misc{liu2024mmdumultiturnmultiimagedialog,
      title={MMDU: A Multi-Turn Multi-Image Dialog Understanding Benchmark and Instruction-Tuning Dataset for LVLMs}, 
      author={Ziyu Liu and Tao Chu and Yuhang Zang and Xilin Wei and Xiaoyi Dong and Pan Zhang and Zijian Liang and Yuanjun Xiong and Yu Qiao and Dahua Lin and Jiaqi Wang},
      year={2024},
      eprint={2406.11833},
      archivePrefix={arXiv},
      primaryClass={cs.CV},
      url={https://arxiv.org/abs/2406.11833}, 
}

@misc{lyu2025vischainbenchbenchmarkmultiturnmultiimage,
      title={VisChainBench: A Benchmark for Multi-Turn, Multi-Image Visual Reasoning Beyond Language Priors}, 
      author={Wenbo Lyu and Yingjun Du and Jinglin Zhao and Xianton Zhen and Ling Shao},
      year={2025},
      eprint={2512.06759},
      archivePrefix={arXiv},
      primaryClass={cs.CV},
      url={https://arxiv.org/abs/2512.06759}, 
}

@misc{tao2025mmsearchplusbenchmarkingprovenanceawaresearch,
      title={MMSearch-Plus: Benchmarking Provenance-Aware Search for Multimodal Browsing Agents}, 
      author={Xijia Tao and Yihua Teng and Xinxing Su and Xinyu Fu and Jihao Wu and Chaofan Tao and Ziru Liu and Haoli Bai and Rui Liu and Lingpeng Kong},
      year={2025},
      eprint={2508.21475},
      archivePrefix={arXiv},
      primaryClass={cs.AI},
      url={https://arxiv.org/abs/2508.21475}, 
}

@misc{shabtay2025livexivmultimodallive,
      title={LiveXiv -- A Multi-Modal Live Benchmark Based on Arxiv Papers Content}, 
      author={Nimrod Shabtay and Felipe Maia Polo and Sivan Doveh and Wei Lin and M. Jehanzeb Mirza and Leshem Chosen and Mikhail Yurochkin and Yuekai Sun and Assaf Arbelle and Leonid Karlinsky and Raja Giryes},
      year={2025},
      eprint={2410.10783},
      archivePrefix={arXiv},
      primaryClass={cs.CV},
      url={https://arxiv.org/abs/2410.10783}, 
}

@misc{wang2024charxivchartinggapsrealistic,
      title={CharXiv: Charting Gaps in Realistic Chart Understanding in Multimodal LLMs}, 
      author={Zirui Wang and Mengzhou Xia and Luxi He and Howard Chen and Yitao Liu and Richard Zhu and Kaiqu Liang and Xindi Wu and Haotian Liu and Sadhika Malladi and Alexis Chevalier and Sanjeev Arora and Danqi Chen},
      year={2024},
      eprint={2406.18521},
      archivePrefix={arXiv},
      primaryClass={cs.CL},
      url={https://arxiv.org/abs/2406.18521}, 
}

@misc{pramanick2025spiqadatasetmultimodalquestion,
      title={SPIQA: A Dataset for Multimodal Question Answering on Scientific Papers}, 
      author={Shraman Pramanick and Rama Chellappa and Subhashini Venugopalan},
      year={2025},
      eprint={2407.09413},
      archivePrefix={arXiv},
      primaryClass={cs.CL},
      url={https://arxiv.org/abs/2407.09413}, 
}

@misc{movva2025enhancingscientificvisualquestion,
      title={Enhancing Scientific Visual Question Answering through Multimodal Reasoning and Ensemble Modeling}, 
      author={Prahitha Movva and Naga Harshita Marupaka},
      year={2025},
      eprint={2507.06183},
      archivePrefix={arXiv},
      primaryClass={cs.CV},
      url={https://arxiv.org/abs/2507.06183}, 
}

@misc{li2023scigraphqa,
  title={SciGraphQA: A Large-Scale Synthetic Multi-Turn Question-Answering Dataset for Scientific Graphs}, 
  author={Shengzhi Li and Nima Tajbakhsh},
  year={2023},
  eprint={2308.03349},
  archivePrefix={arXiv},
  primaryClass={cs.CL}
}

@misc{ren2026sinbenchtracingnativeevidence,
      title={SIN-Bench: Tracing Native Evidence Chains in Long-Context Multimodal Scientific Interleaved Literature}, 
      author={Yiming Ren and Junjie Wang and Yuxin Meng and Yihang Shi and Zhiqiang Lin and Ruihang Chu and Yiran Xu and Ziming Li and Yunfei Zhao and Zihan Wang and Yu Qiao and Ruiming Tang and Minghao Liu and Yujiu Yang},
      year={2026},
      eprint={2601.10108},
      archivePrefix={arXiv},
      primaryClass={cs.CL},
      url={https://arxiv.org/abs/2601.10108}, 
}

@misc{bai2025qwen3vltechnicalreport,
      title={Qwen3-VL Technical Report}, 
      author={Shuai Bai and Yuxuan Cai and Ruizhe Chen and others},
      year={2025},
      eprint={2511.21631},
      archivePrefix={arXiv},
      primaryClass={cs.CV},
      url={https://arxiv.org/abs/2511.21631}, 
}

@misc{vteam2026glm45vglm41vthinkingversatilemultimodal,
      title={GLM-4.5V and GLM-4.1V-Thinking: Towards Versatile Multimodal Reasoning with Scalable Reinforcement Learning}, 
      author={V Team and Wenyi Hong and Wenmeng Yu and others},
      year={2026},
      eprint={2507.01006},
      archivePrefix={arXiv},
      primaryClass={cs.CV},
      url={https://arxiv.org/abs/2507.01006}, 
}

@misc{kimiteam2026kimik25visualagentic,
      title={Kimi K2.5: Visual Agentic Intelligence}, 
      author={Kimi Team and Tongtong Bai and Yifan Bai and others},
      year={2026},
      eprint={2602.02276},
      archivePrefix={arXiv},
      primaryClass={cs.CL},
      url={https://arxiv.org/abs/2602.02276}, 
}

@misc{singh2025openaigpt5card,
      title={OpenAI GPT-5 System Card}, 
      author={Aaditya Singh and Adam Fry and Adam Perelman and others},
      year={2025},
      eprint={2601.03267},
      archivePrefix={arXiv},
      primaryClass={cs.CL},
      url={https://arxiv.org/abs/2601.03267}, 
}

@misc{openai_gpt5_2_2025,
  author       = {{OpenAI}},
  title        = {Introducing GPT-5.2},
  year         = {2025},
  howpublished = {\url{https://openai.com/index/introducing-gpt-5-2/}},
  note         = {Accessed: 2026-03-05}
}

@misc{xai_grok41_2025,
  author       = {{xAI}},
  title        = {Grok 4.1},
  year         = {2025},
  month        = nov,
  howpublished = {\url{https://x.ai/news/grok-4-1}},
  note         = {Accessed: 2026-03-05}
}

@misc{comanici2025gemini25pushingfrontier,
      title={Gemini 2.5: Pushing the Frontier with Advanced Reasoning, Multimodality, Long Context, and Next Generation Agentic Capabilities}, 
      author={Gheorghe Comanici and Eric Bieber and Mike Schaekermann and Ice Pasupat and Noveen Sachdeva and Inderjit Dhillon and Marcel Blistein and Ori Ram and Dan Zhang and Wesley Helmholz},
      year={2025},
      eprint={2507.06261},
      archivePrefix={arXiv},
      primaryClass={cs.CL},
      url={https://arxiv.org/abs/2507.06261}, 
}

@misc{lee2025multiversemultiturnconversationbenchmark,
      title={MultiVerse: A Multi-Turn Conversation Benchmark for Evaluating Large Vision and Language Models}, 
      author={Young-Jun Lee and Byung-Kwan Lee and Jianshu Zhang and Yechan Hwang and Byungsoo Ko and Han-Gyu Kim and Dongyu Yao and Xuankun Rong and Eojin Joo and Seung-Ho Han and Bowon Ko and Ho-Jin Choi},
      year={2025},
      eprint={2510.16641},
      archivePrefix={arXiv},
      primaryClass={cs.CV},
      url={https://arxiv.org/abs/2510.16641}, 
}

@misc{jiang2025memoryqaansweringrecallquestions,
      title={Memory-QA: Answering Recall Questions Based on Multimodal Memories}, 
      author={Hongda Jiang and Xinyuan Zhang and Siddhant Garg and Rishab Arora and Shiun-Zu Kuo and Jiayang Xu and Ankur Bansal and Christopher Brossman and Yue Liu and Aaron Colak and Ahmed Aly and Anuj Kumar and Xin Luna Dong},
      year={2025},
      eprint={2509.18436},
      archivePrefix={arXiv},
      primaryClass={cs.AI},
      url={https://arxiv.org/abs/2509.18436}, 
}

@misc{chen2026efficientmultimodalplanningagent,
      title={Efficient Multimodal Planning Agent for Visual Question-Answering}, 
      author={Zhuo Chen and Xinyu Geng and Xinyu Wang and Yong Jiang and Zhen Zhang and Pengjun Xie and Kewei Tu},
      year={2026},
      eprint={2601.20676},
      archivePrefix={arXiv},
      primaryClass={cs.CL},
      url={https://arxiv.org/abs/2601.20676}, 
}

@misc{wang2024mineruopensourcesolutionprecise,
      title={MinerU: An Open-Source Solution for Precise Document Content Extraction}, 
      author={Bin Wang and Chao Xu and Xiaomeng Zhao and Linke Ouyang and Fan Wu and Zhiyuan Zhao and Rui Xu and Kaiwen Liu and Yuan Qu and Fukai Shang and Bo Zhang and Liqun Wei and Zhihao Sui and Wei Li and Botian Shi and Yu Qiao and Dahua Lin and Conghui He},
      year={2024},
      eprint={2409.18839},
      archivePrefix={arXiv},
      primaryClass={cs.CV},
      url={https://arxiv.org/abs/2409.18839}, 
}

\clearpage
\appendix

\renewcommand{\theHsection}{appendix.\Alph{section}}
\renewcommand{\theHsubsection}{appendix.\Alph{section}.\arabic{subsection}}
\phantomsection
\section*{Supplementary Material}
\label{sec:supplementary}

This supplementary document provides additional details on benchmark construction, evaluation reliability, representative failure cases, and reproducibility for \textsc{EpiBench}. The main paper is self-contained, while this document supplies extended evidence and implementation details.

\noindent\textbf{Contents}
\begin{itemize}
    \item \hyperref[app:const111ruction]{\textbf{A. Benchmark Construction and Composition}}
    \begin{itemize}
        \item \hyperref[app:curation]{A.1 Human Curation and Quality Control}
        \item \hyperref[app:benchmark_composition]{A.2 Benchmark Composition}
    \end{itemize}

    \item \hyperref[app:evaluation_reliability]{\textbf{B. Evaluation Reliability}}

    \item \hyperref[app:failure_cases]{\textbf{C. Representative Failure Cases}}

    \item \hyperref[app:reproducibility]{\textbf{D. Reproducibility Details}}
    \begin{itemize}
        \item \hyperref[app:tool_spec]{D.1 Detailed Tool Specification}
        \item \hyperref[app:efficiency]{D.2 Efficiency and Runtime Statistics}
        \item \hyperref[app:evaluation_prompt]{D.3 Evaluation Prompt}
    \end{itemize}
\end{itemize}

\clearpage
\phantomsection

\section{Benchmark Construction and Composition}
\label{app:const111ruction}
\phantomsection

\subsection{Human Curation and Quality Control}
\label{app:curation}

\textsc{EpiBench} is not released from raw model drafts. Instead, all candidate episodes are refined, validated, and consolidated by five annotators with Ph.D.\ degrees in computer science. Starting from GPT-5.2-generated drafts, the annotators remove ambiguous cases, correct evidence targets, verify that the final turn truly depends on earlier multimodal evidence, and ensure that each required paper contributes essential support to the final answer. Disagreements across annotators are resolved through discussion, followed by iterative refinement of the episode design, including question reformulation and evidence-target revision, until consensus is reached.

\begin{figure}[t]
    \centering
    \includegraphics[width=\linewidth]{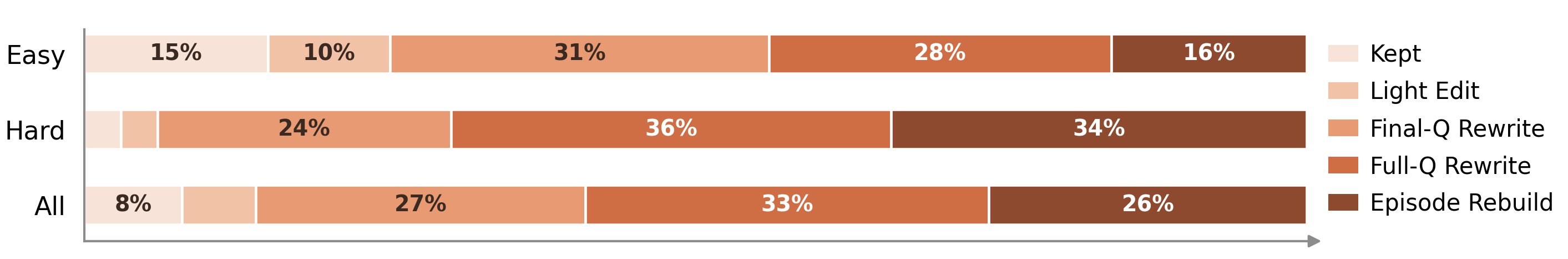}
    \caption{Human revision intensity from GPT-5.2 drafts to the final benchmark release. We group episodes into five categories, ranging from minimal intervention (\textit{Kept}, \textit{Light Edit}) to substantial restructuring (\textit{Final-Q Rewrite}, \textit{Full-Q Rewrite}, \textit{Episode Rebuild}). Most episodes, especially in the hard split, undergo major expert revision before inclusion in \textsc{EpiBench}.}
    \label{fig:revision_breakdown}
\end{figure}

Figure~\ref{fig:revision_breakdown} quantifies the extent of human intervention from the initial GPT-5.2 drafts to the final benchmark release. We categorize revision intensity into five levels: \textit{Kept}, \textit{Light Edit}, \textit{Final-Q Rewrite}, \textit{Full-Q Rewrite}, and \textit{Episode Rebuild}. The benchmark is dominated by substantial human revision. On the full set, only 8\% of episodes are kept as drafted and 6\% receive only light edits, such as wording adjustments, minor question reformulation, or correction of the gold answer. While 27\% require rewriting the final question, 33\% require rewriting all questions in the episode, and 26\% require rebuilding the entire episode, including the associated papers and evidence structure. This trend is even stronger on the hard split, where 94\% of episodes undergo substantial rewriting or rebuilding.

These statistics indicate that the final benchmark is not a lightly post-edited GPT-5.2 product. Rather, its wording, evidence structure, and task constraints are materially shaped by expert human revision. This substantial divergence from the initial drafts helps mitigate concerns that the benchmark construction process is overly tied to GPT-5.2-specific phrasing or reasoning patterns.

\phantomsection

\subsection{Benchmark Composition}
\label{app:benchmark_composition}

\begin{figure*}[t]
    \centering
    \begin{minipage}[t]{0.38\textwidth}
        \centering
        \includegraphics[width=\linewidth]{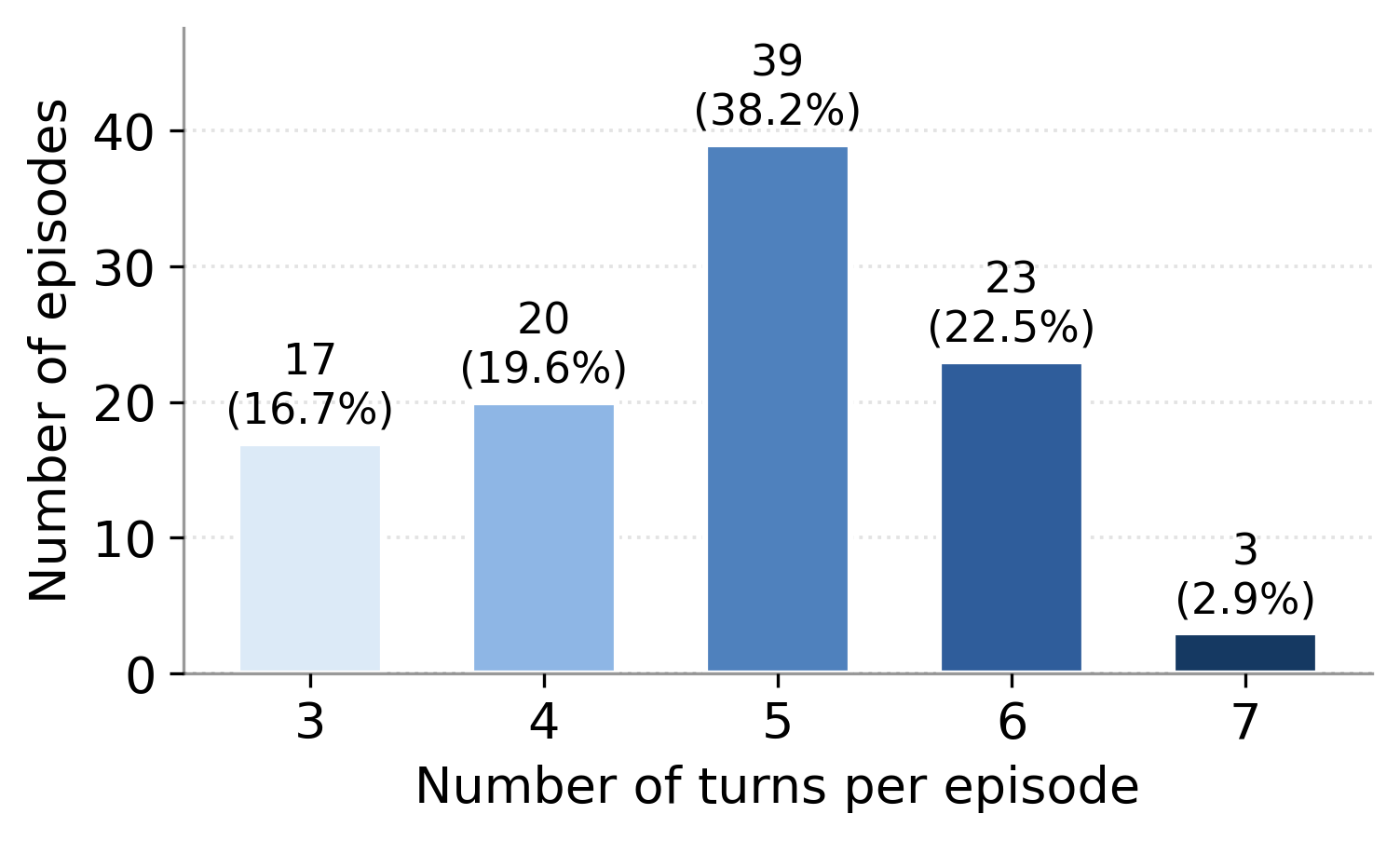}
    \end{minipage}
    \hfill
    \begin{minipage}[t]{0.29\textwidth}
        \centering
        \includegraphics[width=\linewidth]{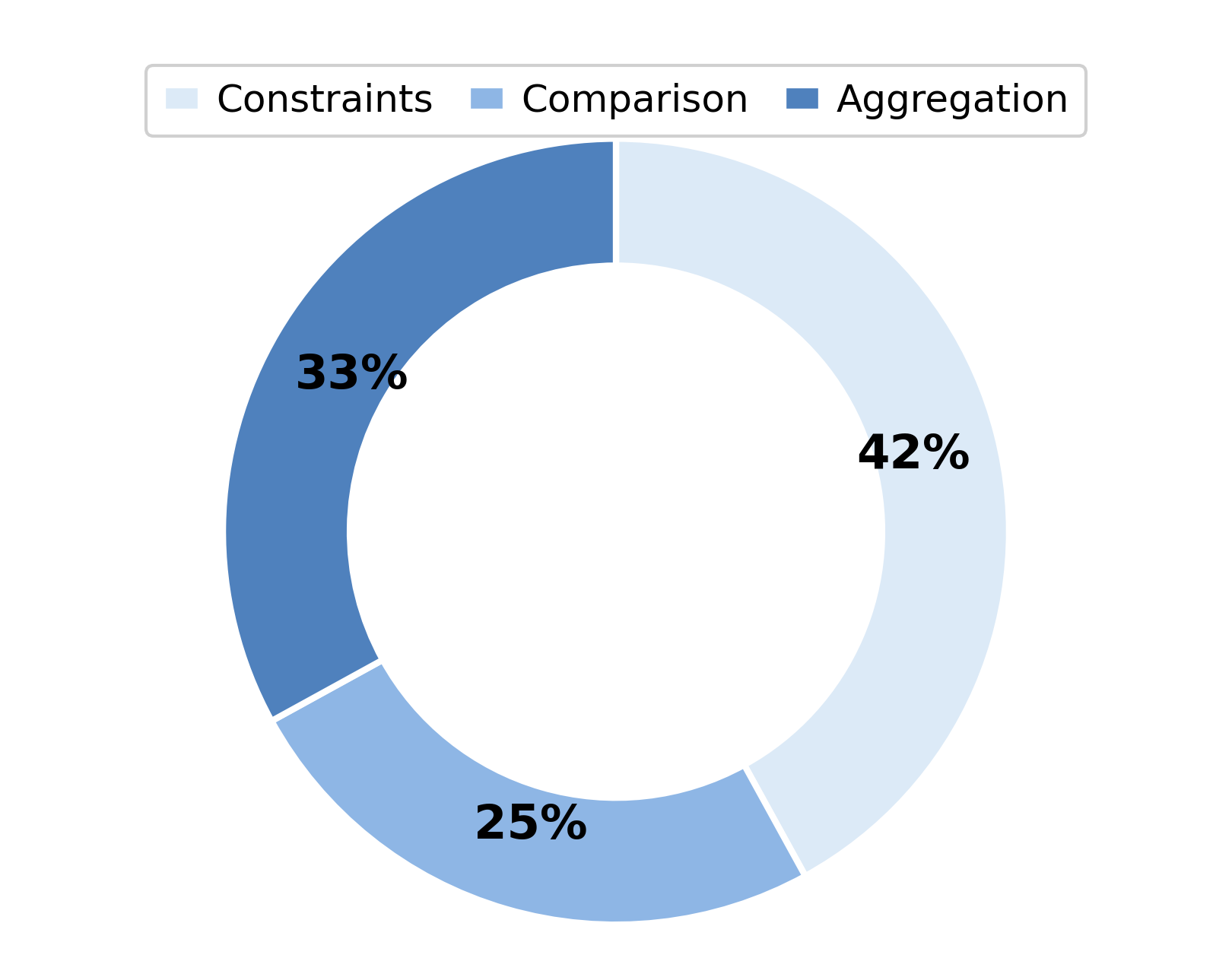}
    \end{minipage}
    \hfill
    \begin{minipage}[t]{0.29\textwidth}
        \centering
        \includegraphics[width=\linewidth]{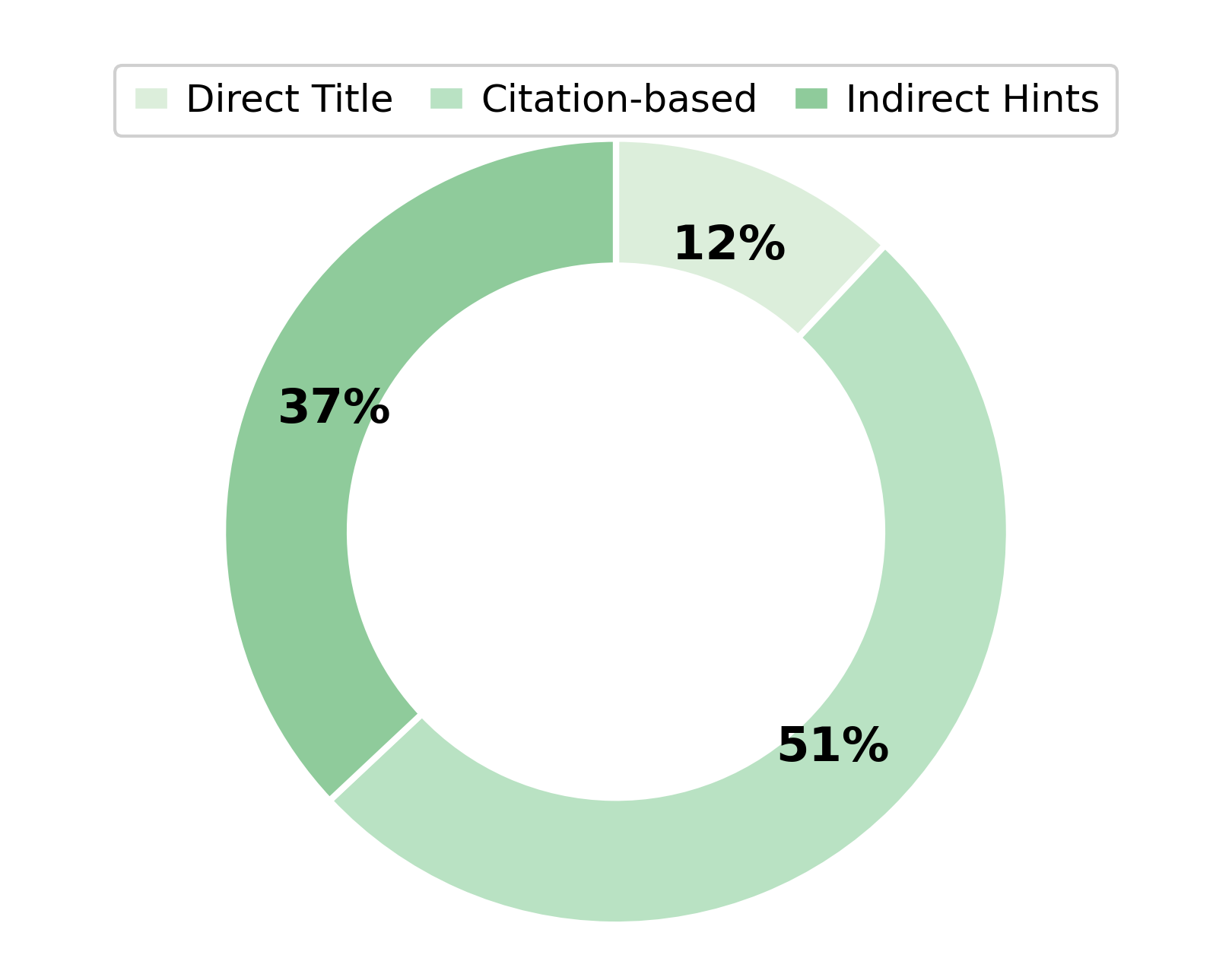}
    \end{minipage}
    \caption{Statistics of benchmark composition. Left: distribution of episode turn lengths. Middle: distribution of task categories, including constraint-style identification, cross-paper comparison, and evidence aggregation. Right: distribution of proactive search initiation types, including direct-title search, citation-based search, and indirect-hint-driven search.}
    \label{fig:benchmark_composition}
\end{figure*}

Figure~\ref{fig:benchmark_composition} summarizes the composition of \textsc{EpiBench} from three perspectives: episode turn length, task categories, and proactive search initiation types.

Left, we report the distribution of episode turn lengths. Most episodes contain between 4 and 6 turns, with 5-turn episodes being the most common. This reflects our design goal of modeling short but non-trivial research workflows that require multi-step evidence acquisition and integration, while remaining compact enough for controlled evaluation.

Middle, we categorize benchmark questions into three task types. \textit{Constraint-style} tasks require agents to identify papers or settings that satisfy a set of conditions, such as matching multiple experimental requirements or filtering candidates under joint constraints. \textit{Comparison} tasks require explicit cross-paper comparison over methods, settings, or reported results. \textit{Aggregation} tasks require agents to collect and combine evidence from multiple papers, figures, or tables before producing the final answer. This distribution shows that the benchmark is not centered on a single answer pattern, but instead covers several common forms of multi-step research reasoning.

Right, we categorize episodes by how proactive search is initiated. \textit{Direct title} cases provide an explicit paper title and mainly test retrieval and evidence extraction after the target is known. \textit{Citation-based} cases provide citation cues, such as author names, venue information, or referenced relations, requiring agents to locate the target paper through bibliographic clues. \textit{Indirect hint} cases provide only partial semantic descriptions, task context, or experimental characteristics, without direct identifiers, requiring more open-ended literature search. The resulting distribution shows that a substantial portion of \textsc{EpiBench} requires proactive discovery rather than direct lookup, which is consistent with our workflow-faithful benchmark design.

\phantomsection

\section{Evaluation Reliability}
\label{app:evaluation_reliability}

To assess whether the main evaluation is overly dependent on the GPT-5.2-based judge, we conduct a reliability study on a sampled subset covering 30\% of the predictions from all eight models in the main experiments. We invite an independent model judge (Gemini-2.5-Pro) and two Ph.D.-level human judges to re-evaluate the same outputs under the same rubric, and compare their decisions with those of the primary GPT-5.2 judge. The human judges only see the task, the model output, and the scoring rubric, but not the original GPT-5.2-judge decision. We report agreement rates and Cohen's $\kappa$ at both the turn and episode levels. Let $y_i^{(a)}\in\{0,1\}$ and $y_i^{(b)}\in\{0,1\}$ denote the binary decisions made by judges $a$ and $b$ on sample $i$. The agreement rate is computed as
\begin{equation}
\mathrm{Agr}(a,b)=\frac{1}{M}\sum_{i=1}^{M}\mathbb{I}\!\left[y_i^{(a)}=y_i^{(b)}\right],
\end{equation}
where $M$ is the number of evaluated samples and $\mathbb{I}[\cdot]$ is the indicator function. To discount chance agreement, we additionally report Cohen's $\kappa$:
\begin{equation}
\kappa(a,b)=\frac{p_o-p_e}{1-p_e},
\end{equation}
where $p_o=\mathrm{Agr}(a,b)$ is the observed agreement and $p_e$ is the expected agreement under the empirical label marginals of the two judges. As shown in Table~\ref{tab:judge_agreement}, all judge pairs exhibit very high agreement, which suggests that the main evaluation is stable and not materially tied to a single judge model.

\begin{table}[t]
\caption{Agreement analysis between the primary GPT-5.2-based judge, an independent Gemini-2.5-Pro judge, and Ph.D.-level human judges on a sampled subset covering 30\% of the predictions from all eight models in the main experiments. Higher is better for all metrics.}
\label{tab:judge_agreement}
\centering
\footnotesize
\setlength{\tabcolsep}{4pt}
\renewcommand{\arraystretch}{1.12}
\begin{tabular}{lcccc}
\toprule
Judge Pair & Turn Agr. $\uparrow$ & Turn $\kappa$ $\uparrow$ & Episode Agr. $\uparrow$ & Episode $\kappa$ $\uparrow$ \\
\midrule
GPT vs. Gemini & 98.7 & 0.97 & 100.0 & 1.00 \\
GPT vs. Human  & 100.0 & 1.00 & 100.0 & 1.00 \\
Human vs. Human & 100.0 & 1.00 & 100.0 & 1.00 \\
\bottomrule
\end{tabular}
\end{table}

\phantomsection

\section{Representative Failure Cases}
\label{app:failure_cases}

\begin{figure*}[t]
    \centering
    \includegraphics[width=\textwidth]{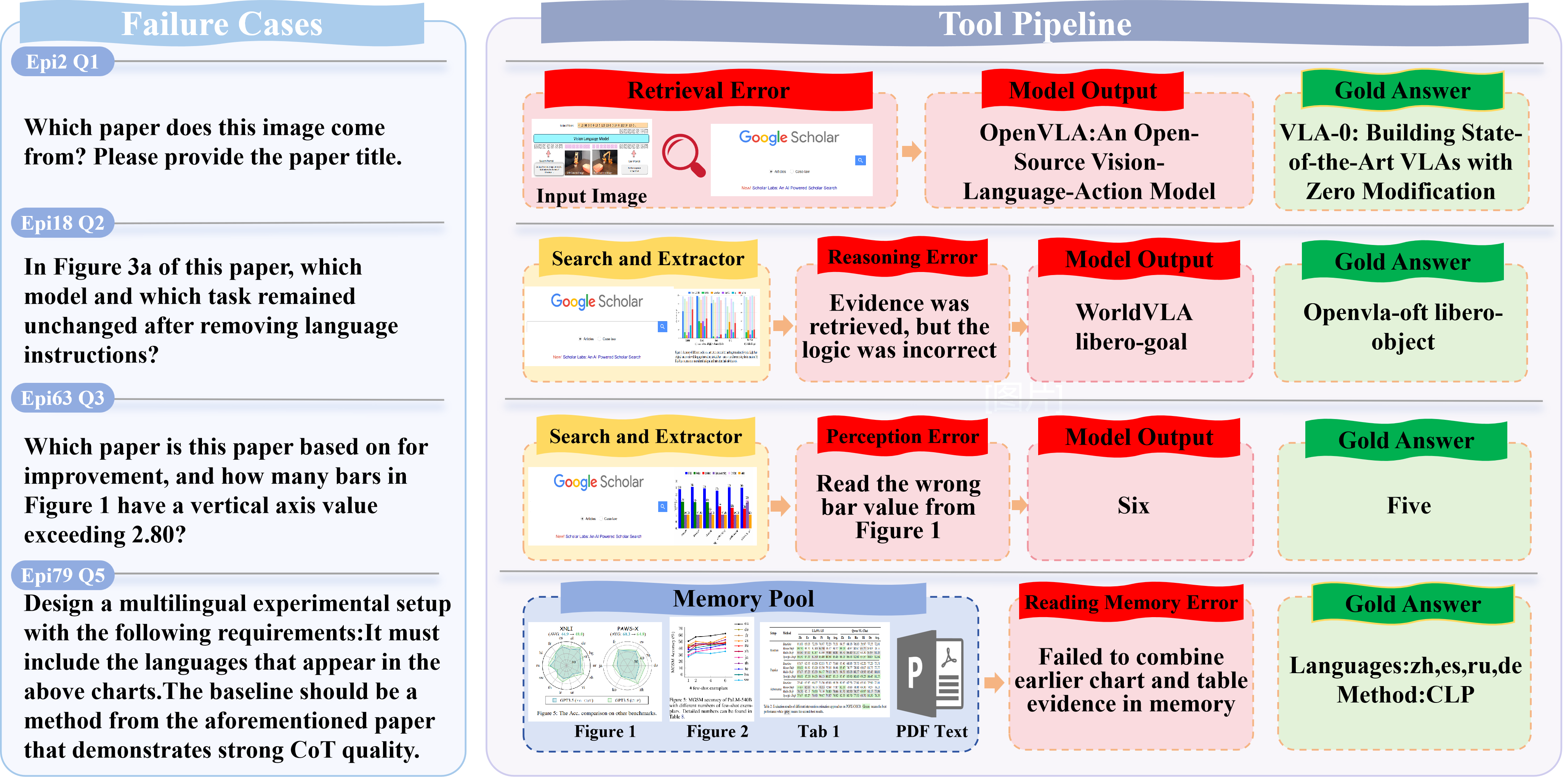}
    \caption{Representative failed questions grouped by error type. Each row shows a compressed execution trace from a real failed episode. The highlighted red module marks the first error under our trace-based attribution rule, while later incorrect outputs are shown only as downstream consequences. From top to bottom, the examples illustrate retrieval failure, reasoning failure, perception failure, and reading-memory failure. The omitted \textit{Others} category mainly includes max-step termination, tool execution failures, and runtime or parsing issues.}
    \label{fig:error_case_examples}
\end{figure*}

Figure~\ref{fig:error_case_examples} presents representative failed questions for the main error categories used in our trace-based analysis. Each row shows a compressed execution trace from a failed episode. The highlighted red module marks the first error under our attribution rule, while later incorrect outputs are shown only as downstream consequences for context. This visualization complements the aggregate error statistics in the main paper by illustrating how different failure types arise in concrete benchmark interactions.

We categorize failures into five types. \textbf{Retrieval} errors occur when the agent fails to identify or access the correct paper, typically due to misleading search cues, incorrect paper matching, or unsuccessful search-followup behavior. \textbf{Perception} errors occur when the correct evidence source has already been reached, but the agent misreads the content of a target figure or table, such as values, labels, or row--column correspondences. \textbf{Reasoning} errors occur when the relevant evidence has been successfully retrieved and correctly read, but the agent fails to derive the answer through comparison, constraint composition, or multi-evidence aggregation. \textbf{Reading Memory} errors occur when previously collected evidence exists in memory, but the agent fails to retrieve, align, or reuse it correctly in a later turn, especially in the final no-tool setting. \textbf{Others} covers residual non-core cases, including max-step termination, tool execution failures, and runtime or parsing issues. We omit this category from the main analysis because it is rare, accounting for only around 3\% of all errors, and is not directly reflective of core model capability.

\phantomsection

\section{Reproducibility Details}
\label{app:reproducibility}
\phantomsection

\subsection{Detailed Tool Specification}

\label{app:tool_spec}

The default \textsc{EpiBench} protocol in the main text uses four \emph{core} tools for literature discovery and paper content access, namely \textsc{Web Search}, \textsc{PDF Extractor}, \textsc{PDF Extractor RAG}, and \textsc{Content Extractor}. To make the implementation more explicit in the appendix, we further decompose \textsc{Content Extractor} into two concrete specializations, \textsc{Figure Extractor} and \textsc{Table Extractor}, because they operate on different evidence types and return different structured outputs. In addition, we report an auxiliary \textsc{Code Executor} tool for completeness. This tool is listed separately because it is not part of the core reasoning process of the agent, and instead serves only as a simple execution utility for code-based operations.

All tools follow a unified interface and return structured outputs that can be appended to memory and traced during evaluation. Table~\ref{tab:tool_spec} summarizes their roles, inputs, and outputs.


\newcommand{\toolcell}[1]{\makecell[tl]{\bfseries\scshape #1}}

\begin{table*}[t]
\caption{Detailed specification of the tool environment. The first four entries correspond to the core benchmark toolchain used in the main text, with \textsc{Figure Extractor} and \textsc{Table Extractor} serving as explicit decompositions of \textsc{Content Extractor}. \textsc{Code Executor} is listed separately because it is not part of the agent's core reasoning process, but only a simple auxiliary execution utility, and is therefore not included in the default benchmark protocol.}
\label{tab:tool_spec}
\centering
\scriptsize
\setlength{\tabcolsep}{4pt}
\renewcommand{\arraystretch}{1.3}
\begin{tabularx}{\textwidth}{
>{\raggedright\arraybackslash}p{1.45cm}
>{\raggedright\arraybackslash}X
>{\raggedright\arraybackslash}p{3.0cm}
>{\raggedright\arraybackslash}p{3.0cm}
}
\toprule
\textbf{Tool} & \textbf{Function} & \textbf{Input} & \textbf{Output} \\
\midrule

\toolcell{Web\\\textbf{Search}}
& Discovers candidate papers, resolves paper identities from indirect cues, and follows citation leads through ranked web retrieval.
& Natural-language query, citation cue, keyword phrase
& Ranked result list with titles, snippets, and URLs \\

\toolcell{PDF\\\textbf{Extractor}}
& Downloads a paper PDF and parses it into a normalized cached representation for later access.
& Paper identifier such as title, URL, or arXiv ID
& Parsed markdown document, metadata, and cache handle \\

\toolcell{PDF\\\textbf{Extractor}\\\textbf{RAG}}
& Retrieves the most relevant text spans from a cached paper without rereading the entire document.
& Paper identifier plus text query
& Ranked text spans with section context and lightweight provenance \\

\toolcell{Figure\\\textbf{Extractor}}
& Extracts a target figure from a cached paper when visual grounding is required.
& Paper identifier plus figure index or figure ID
& Figure image, caption context, nearby text, and provenance \\

\toolcell{Table\\\textbf{Extractor}}
& Extracts a target table from a cached paper and returns a structured representation for downstream reasoning.
& Paper identifier plus table index or table ID
& Table content, structured markdown, nearby text, and provenance \\

\toolcell{Code\\\textbf{Executor}}
& Executes code for deterministic calculations, lightweight post-processing, or verification in extended settings.
& Code snippet with optional serialized inputs
& Execution result, derived values, and runtime logs such as stdout or stderr \\

\bottomrule
\end{tabularx}
\end{table*}

\phantomsection

\subsection{Efficiency and Runtime Statistics}
\label{app:efficiency}

Table~\ref{tab:efficiency_stats} reports aggregate token usage and runtime statistics over representative full-benchmark runs. A notable finding is the substantial time gap between human experts and model-based agents. The Ph.D.\ expert baseline requires 14.04 in units of $\times 10^{4}$ seconds to complete the benchmark, whereas model runtimes range from only 2.64 to 9.45. This result highlights an important property of our benchmark: although current agents still fall far short of expert-level success, they can already execute long-horizon research workflows at a much lower time cost than human workers. Therefore, developing more capable and reliable agents for this type of workflow is a highly valuable research direction, as even moderate improvements in task success could translate into substantial reductions in human labor and end-to-end completion time.

Token usage also differs substantially across models. In particular, some agents still require very large token budgets despite relatively low task success. For example, Qwen3-VL-235B-Instruct reaches 5.67 in units of $\times 10^{7}$ total tokens, indicating that current research agents remain costly even when their reliability is far from satisfactory. This suggests that better balancing agent performance and interaction cost remains an open question in current research on autonomous research agents.

\begin{table}[t]
\caption{Aggregate token usage and runtime statistics over representative full-benchmark runs. Input and total tokens are reported in units of $\times 10^{7}$, output tokens are reported in units of $\times 10^{6}$, and duration is reported in units of $\times 10^{4}$ s.}
\label{tab:efficiency_stats}
\centering
\scriptsize
\setlength{\tabcolsep}{4pt}
\renewcommand{\arraystretch}{1.12}
\begin{tabular}{lcccc}
\toprule
Model & Input ($\times 10^{7}$) & Output ($\times 10^{6}$) & Total ($\times 10^{7}$) & Duration ($\times 10^{4}$ s) \\
\midrule
Kimi-K2.5              & 3.58 & 1.87 & 3.77 & 5.68 \\
GLM-4.5V               & 1.52 & 1.04 & 1.62 & 3.61 \\
Grok-4.1          & 3.99 & 2.08 & 4.20 & 3.38 \\
Gemini-2.5-Pro         & 3.49 & 1.08 & 3.60 & 2.64 \\
GPT-5.2                & 2.10 & 2.06 & 2.31 & 3.64 \\
GPT-5-Mini             & 3.25 & 1.43 & 3.39 & 3.52 \\
Qwen3-VL-235B-Instruct & 5.54 & 1.23 & 5.67 & 9.45 \\
Qwen3-VL-235B-Thinking & 1.02 & 1.33 & 1.16 & 3.06 \\
\midrule
Ph.D.\ Experts         & --   & --   & --   & 14.04 \\
\bottomrule
\end{tabular}
\end{table}

\phantomsection

\subsection{Evaluation Prompt}
\label{app:evaluation_prompt}

We report below the instruction template used to prompt model-based agents during evaluation. A shared base prompt is used for all turns, with additional suffixes appended for choice questions, multi-turn choice questions, and the final no-tool turn.

\begin{promptbox}{Prompt Used for Model-Based Agents}
[Base System Instruction]

You are a research agent that answers questions by reading PDFs.

[Planning and Execution]
- First, review the conversation history and memory to determine whether the task has already been completed or whether relevant information, such as paper identifiers or previously extracted content, is already available.
- If the required information is already in memory, use it directly instead of retrieving it again.
- Plan tool calls efficiently. Retrieve information once, then analyze it thoroughly.
- Avoid redundant operations. Do not search for the same target repeatedly.

[Workflow]
1. Identify what information is needed and check whether it is already available.
2. If it is not available, retrieve it efficiently through search, download, or extraction tools.
3. Analyze the collected evidence and answer the question based on the retrieved sources.

[Best Practices]
- Do not rely on prior knowledge. Use tools to retrieve evidence.
- Do not call final_answer in the same response as any other tool. It should be used alone only after all reasoning and tool calls are complete.
- If the paper identifier is already known from the context, use it directly.
- Extract only one image in a single response, since extracted images are automatically appended to the context.
- Verify the answer against the source evidence before finalizing.

[Instruction for Choice Questions]
This is a choice question. The answer may contain a single option or multiple options. Choose only from the provided options. If multiple options are correct, concatenate them into a single string. For example, if the correct choices are A, B, and C, the final answer should be "ABC".

[Instruction for Multi-Turn Choice Questions]
This question contains several multi-turn choice sub-questions. For each sub-question, choose from the provided options and concatenate the selected choices into a single string. Separate the answers to different sub-questions with a semicolon. For example, if the correct answer is "ABC" for the first sub-question and "AC" for the second, the final answer should be "ABC; AC".

[Instruction for the Final No-Tool Turn]
This is the final turn. Do not call any tool. Carefully review the previous reasoning steps and accumulated evidence in memory, then provide the final answer directly in the following format:

Thought:<your thoughts>
<code>final_answer('...')</code>
\end{promptbox}

\end{document}